\documentclass[journal]{IEEEtran}

\usepackage{graphicx}
\usepackage{float}
\usepackage{amsthm,amsmath,amssymb}
\usepackage{mathrsfs}
\usepackage{dutchcal}
\usepackage{multirow}
\usepackage{multicol}
\usepackage{setspace}
\usepackage{caption}
\usepackage{cite}
\usepackage{color}
\usepackage[colorlinks,linkcolor=blue,citecolor=blue]{hyperref}
\usepackage{makecell}
\usepackage{array}
\usepackage{booktabs}
\usepackage{colortbl}
\usepackage{color}
\usepackage{graphicx}
\usepackage{subfig}

\newcommand{\ie}{\textit{i}.\textit{e}.}
\newcommand{\eg}{\textit{e}.\textit{g}.}

\usepackage{bbm}

\begin{document}
%
\title{Query-guided Prototype Evolution Network for Few-Shot Segmentation}


\author{Runmin Cong,~\IEEEmembership{Senior Member,~IEEE}, Hang Xiong, Jinpeng Chen,~\IEEEmembership{Student Member,~IEEE}, Wei Zhang,~\IEEEmembership{Senior Member,~IEEE}, Qingming Huang,~\IEEEmembership{Fellow,~IEEE}, and Yao Zhao,~\IEEEmembership{Fellow,~IEEE} 

\thanks{Corresponding author: Jinpeng Chen.}
\thanks{Runmin Cong is with the Institute of Information Science, Beijing Jiaotong University, Beijing 100044, China, also with the School of Control Science and Engineering, Shandong University, Jinan 250061, China, and also with the Key Laboratory of Machine Intelligence and System Control, Ministry of Education, Jinan 250061, China (e-mail: rmcong@sdu.edu.cn).}
\thanks{Hang Xiong and Yao Zhao are with the Institute of Information Science, Beijing Jiaotong University, Beijing 100044, China, also with the Beijing Key Laboratory of Advanced Information Science and Network Technology, Beijing 100044, China (e-mail: xionghang@bjtu.edu.cn; yzhao@bjtu.edu.cn).}
\thanks{Jinpeng Chen is with the Department of Computer Science, City University of Hong Kong, Hong Kong SAR, China (e-mail: jinpechen2-c@my.cityu.edu.hk).}
\thanks{Wei Zhang is with the School of Control Science and Engineering, Shandong University, Jinan 250061, China, and also with the Key Laboratory of Machine Intelligence and System Control, Ministry of Education, Jinan 250061, China (e-mail: davidzhang@sdu.edu.cn).}
\thanks{Qingming Huang is with the School of Computer Science and Technology, University of Chinese Academy of Sciences, Beijing 101408, China.
E-mail: qmhuang@ucas.ac.cn (e-mail: qmhuang@ucas.ac.cn).}

}

%
%

\markboth{IEEE Transactions on Multimedia}%
{Shell \MakeLowercase{\textit{et al.}}: Bare Demo of IEEEtran.cls for IEEE Journals}
%



\maketitle

\begin{abstract}
Previous Few-Shot Segmentation (FSS) approaches exclusively utilize support features for prototype generation, neglecting the specific requirements of the query. To address this, we present the Query-guided Prototype Evolution Network (QPENet), a new method that integrates query features into the generation process of foreground and background prototypes, thereby yielding customized prototypes attuned to specific queries. 
The evolution of the foreground prototype is accomplished through a \textit{support-query-support} iterative process involving two new modules: Pseudo-prototype Generation (PPG) and Dual Prototype Evolution (DPE). The PPG module employs support features to create an initial prototype for the preliminary segmentation of the query image, resulting in a pseudo-prototype reflecting the unique needs of the current query. Subsequently, the DPE module performs reverse segmentation on support images using this pseudo-prototype, leading to the generation of evolved prototypes, which can be considered as custom solutions. As for the background prototype, the evolution begins with a global background prototype that represents the generalized features of all training images. We also design a Global Background Cleansing (GBC) module to eliminate potential adverse components mirroring the characteristics of the current foreground class. Experimental results on the PASCAL-$5^i$ and COCO-$20^i$ datasets attest to the substantial enhancements achieved by QPENet over prevailing state-of-the-art techniques, underscoring the validity of our ideas.
\end{abstract}


%
\IEEEpeerreviewmaketitle

\section{Introduction}

\begin{figure}[t]
    \centering
    \includegraphics[width=1\linewidth]{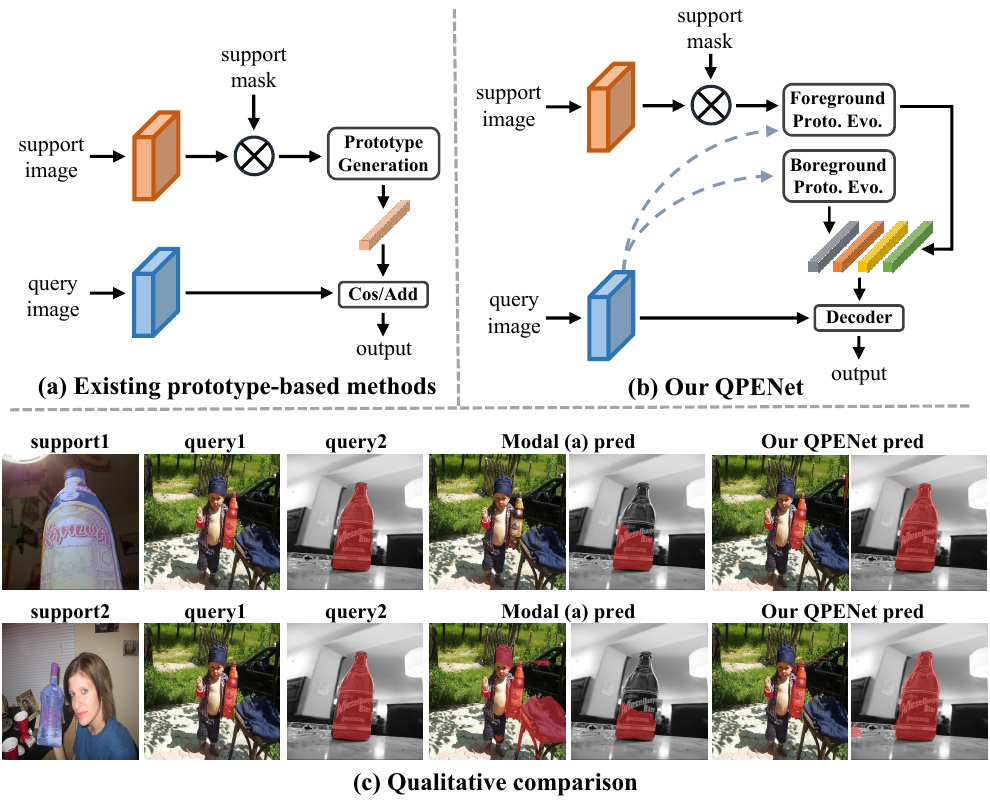}
    \caption{Comparison of prototype generation strategies in existing prototype-based methods and our QPENet. (a) The strategy in existing prototype-based methods that only considers the support image. (b) The strategy in our QPENet that considers both the support and query images. (c) A qualitative comparison of the previous prototype-based method \cite{tian2020prior} and our QPENet. 
    }
    \label{idea}
\end{figure}

Deep learning advancements \cite{DBLP:journals/jei/LiGWCZW16, DBLP:journals/access/NiLCZPF17, crm/tip20/MCMT-GAN, DBLP:journals/tip/LiGRCHKT20, crm/jbhi22/polyp, chen2023saving, crmbridgenet, crm/acmmm23/FPNet, crm/acmmm23/SDDNet, zhang2023controlvideo} are driving substantial improvements in computer performance within the domain of semantic segmentation \cite{crm/tomm22/hippocampus, wang2019learning, kang2018depth, zhan2019unmanned, yao2021non, crm/tmm22/TNet, chen2022saliency, asgari2021deep, krizhevsky2017imagenet, crm/tce22/covid, DBLP:journals/caaitrit/ZhaoZGY23}. 
However, achieving desired performance in most semantic segmentation networks largely depends on adequate training images and meticulous pixel-level annotations\cite{crm/tnnls22/360SOD, DBLP:journals/tmm/YinTYXW22, crm/tcyb23/instanceSOD, crm/ACMMM20/DMVOS, DBLP:journals/tmm/LiuGMLW21, crm/tim22/covid, DBLP:journals/tmm/WangJSCJ16, crm/tcyb22/glnet}, which can be laborious to acquire.
Semi-supervised or weakly-supervised methods\cite{DBLP:journals/tmm/ZhouGLF21, DBLP:journals/tmm/ZhangLCSSK19, crm/ACMMM23/PICRNet, DBLP:journals/caaitrit/ZhengZZLL22, crm/tcsvt22/weaklySOD}, though they reduce annotation workload, still require ample training image data for all categories, posing problems for rarely seen classes in real-world scenarios. On the other hand, humans can leverage their own experience to master how to segment a new category with just a few example images. 
To emulate this human ability, few-shot semantic segmentation (FSS) has been developed. Its task is to precisely segment the regions of an unseen category within the query image, informed solely by a few labeled support images of that category.

Current FSS methods predominantly employ an episodic training paradigm \cite{vinyals2016matching}, which is composed of a series of episodes, each encompassing a query image and a few support images sourced from the training set. This paradigm emulates the learning process for unseen classes in the few-shot scenario, empowering the network with the ability to learn to segment a novel class from just a handful of support images, which embodies the concept known as \textit{learn-to-learn}\cite{zhang2020sg}. 
As a widely adopted idea, the primary workflow of prototype matching based methods \cite{yang2020prototype, liu2020part, li2021adaptive, zhang2021self, liu2022learning, tian2020prior, wang2019panet, xie2021few, xie2021scale, DBLP:journals/tmm/ChenXYWSTZ22, zhang2021rich} can be summarized as exploring the foreground features from the support images to generate prototypes, and then using the prototypes for feature matching to segment the target region in the query image.

Clearly, prototype quality is crucial in prototype-based FSS methods. During prototype generation, previous works either simply conduct masked average pooling on the foreground of the support features \cite{dong2018few, tian2020prior, wang2019panet, xie2021few, xie2021scale} or employ algorithms such as K-means or EM to decompose the support features \cite{liu2020part, yang2020prototype}. Nevertheless, irrespective of the specific procedure, prototype generation typically only consider the support features, as depicted in Fig. \ref{idea}(a). We argue that this strategy is not optimal as it disregards the diverse requirements of distinct query instances. As demonstrated by the examples in Fig. \ref{idea}(c), different support images can lead to noticeable variations in the segmentation quality of the same query image, and the prototype generated from a support image can perform differently when segmenting various query images. To address this, we introduce the Query-guided Prototype Evolution Network (QPENet), a novel approach that integrates the features of both support and query into prototype generation, as shown in Fig. \ref{idea}(b), allowing for the generation of customized prototypes tailored to the needs of different query images.

Typically, the prototype is used to condense the foreground features. However, a major challenge in incorporating query features into the foreground prototype lies in the absence of the ground-truth query mask, complicating the exclusive inclusion of desired foreground parts in the customized foreground prototype. To overcome this obstacle, we design a prototype evolution strategy featuring two critical modules: Pseudo-prototype Generation (PPG) and Dual Prototype Evolution (DPE). 
The process begins with the PPG module generating a preliminary prototype from the support features, enabling an initial coarse segmentation of the query image and producing a pseudo-prototype from the query features.
This pseudo-prototype, which primarily encapsulates the foreground features of the query image, reflects the specific requirements of the current query. 
Following this, the DPE module utilizes the pseudo-prototype to reverse-segment the support image, producing a match between the query's specific requirements and the support features, which can be considered as customized solutions. Throughout this \textit{support-query-support} process, the initial, inflexible prototype evolves into adaptive prototypes tailored for the current query.

In addition to the foreground prototype, some earlier works also incorporated background prototypes to exclude background regions. Typically, background prototypes \cite{lang2022beyond, yang2020prototype} are also based solely on the current support images. The limitation here is that while the foreground class is consistent between query and support images, the background can vary dramatically, resulting in an ineffective background prototype. To address this issue, \cite{liu2022learning} introduced a global background prototype that accumulates background features from all training images, improving the likelihood of matching the query background. However, this strategy may lead to false exclusions since the current foreground class may appear as background in other episodes and be merged into the global background prototype. To deliver this, akin to the evolution of foreground prototypes, we also integrate query features into the background prototype evolution process, allowing for the generation of a tailored background prototype for each query image. Specifically, we propose a Global Background Cleansing (GBC) module, which firstly estimates the background mask of the given query, and then generates a "cleaner" to filter out components that may represent the current foreground class from the global background prototype.

In general, the core idea of this paper is the utilization of the query to guide the evolution of both foreground and background prototypes, thereby creating customized solutions for the segmentation of the current query. To summarize, our contributions have three aspects: 
\begin{itemize}
\item We propose a novel FSS method named QPENet, which embodies the core idea of using the query image to guide the evolution of prototypes, thereby enhancing their efficacy for segmenting the specific query image.
\item We introduce innovative PPG and DPE modules to facilitate the evolution of the foreground prototype following a \textit{support-query-support} process, and a novel GBC module to eliminate components that might reflect the current foreground class from the global background prototype. 
\item Extensive experiments conducted on two widely adopted datasets demonstrate that our model excels in delivering state-of-the-art performance in the domain of FSS.
\end{itemize}

\begin{figure*}[t]
    \center{\includegraphics[width=\textwidth]{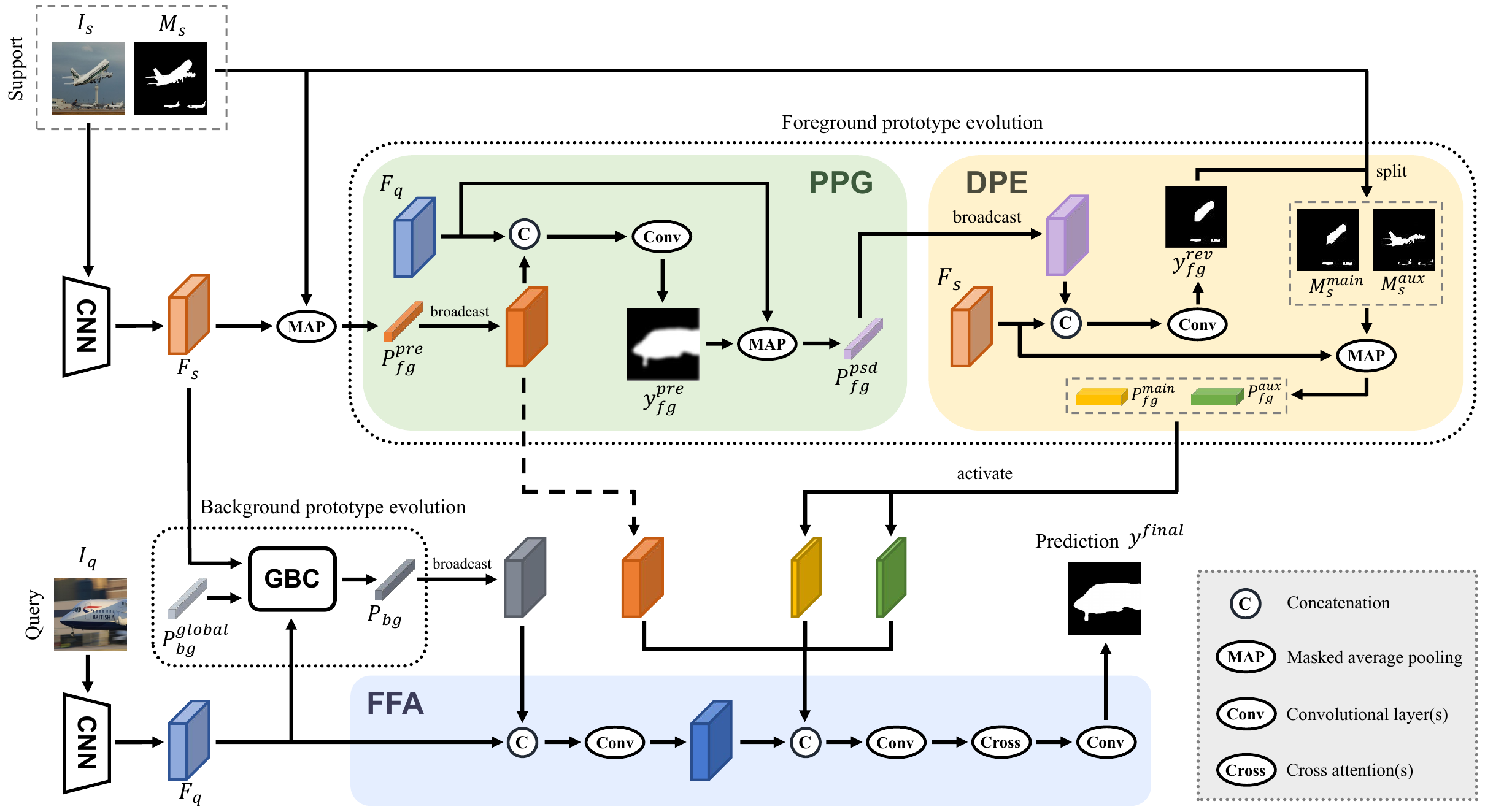}} 
    \caption{Overall architecture of our proposed QPENet.  Initially, it extracts features from both the support and query images using a pre-trained backbone. Subsequently, a foreground prototype evolution process, composed of the PPG and DPE modules, facilitates the creation of a foreground prototype specifically tailored to the current query. Concurrently, a background prototype evolution process, executed by the GBC module, eliminates potential components reflecting the current foreground class from the global background prototype, yielding a customized background prototype. Finally, the FFA module takes the query feature and multiple foreground and background prototypes as input, generating the final prediction.}
    \label{overview}
\end{figure*}

\section{Related Works}

\subsection{Semantic Segmentation}
Semantic segmentation, a pivotal task in computer vision, aims to classify each pixel in an image into distinct object or stuff classes. The emergence of fully convolutional networks (FCNs) \cite{long2015fully} marked a significant leap in this domain. This progress was followed by the introduction of diverse network architectures, components, and attention mechanisms, such as encoder-decoder architectures \cite{chen2018encoder} and feature pyramid modules \cite{zhao2017pyramid, he2019adaptive, crm/tip22/CIRNet}, all aiming at enhancing performance. PSPNet \cite{zhao2017pyramid} and DeepLabV2 \cite{chen2014semantic} integrated multi-scale context between convolution layers. Subsequently, Non-local \cite{wang2018non}, CCNet \cite{huang2019ccnet}, and DGMN \cite{zhang2020dynamic} incorporated the attention mechanism into the convolution structure. With the advent of Vision Transformers, Transformer-based methods were introduced. More recently, MaskFormer \cite{cheng2021per} and Mask2Former \cite{cheng2022masked} achieved semantic segmentation using bipartite matching. Generally, these methods utilized the discriminative pixel-wise classification learning paradigm.

However, these models require numerous images and time-consuming pixel-level annotations to achieve satisfactory results. Meanwhile, it is impossible for them to generalize to unseen categories without fine-tuning, which can be a significant limitation.

\subsection{Few-Shot Segmentation}
 Over the years, the computer vision community has been continuously working on developing networks with the ability to generalize to new categories, leading to the emergence of the Few-Shot Learning (FSL) task \cite{wang2020generalizing, ravi2017optimization, sung2018learning, snell2017prototypical, sun2019meta, oreshkin2018tadam}. The objective of FSL was to train a model capable of proficiently identifying new classes given limited data \cite{allen2019infinite, koch2015siamese, jamal2019task}. The majority of current FSL approaches adhered to the meta-learning framework suggested by \cite{vinyals2016matching}, wherein a set of learning tasks were selected from the base dataset to mimic few-shot scenarios during training \cite{finn2017model, gordon2018meta, grant2018recasting, chen2019image}.

As a subtask of FSL, Few-Shot Segmentation (FSS) has been introduced to address the dense prediction problem in a low-data environment. This involves learning to segment target objects in a query image given a few pixel-wise annotated support images. The initial handling of this task was by OSLSM \cite{shaban2017one}, which employed the support branch to generate classifier weights for prediction in the query branch. Recently, most FSS methods have adopted a metric learning strategy, with prototype-based techniques being prevalent. For instance, \cite{snell2017prototypical} developed prototypes for diverse classes and utilized cosine similarity between the features and prototypes for segmentation. PANet \cite{wang2019panet} integrated prototype alignment regularization to foster consistent embedding prototypes, thus enhancing performance. Further, PFENet \cite{tian2020prior} devised an effective feature pyramid module and utilized a prior map to yield superior segmentation performance. Recent works like PMMs \cite{yang2020prototype} and PPNet \cite{liu2020part} underscored the inadequacy of representing a category with a single support prototype, leading them to use multiple prototypes via EM algorithms or K-means clustering for support object representation. More recent techniques employed multiple prototypes, such as SCL \cite{zhang2021self}, which used an auxiliary support vector to reinforce easily lost information. ASGNet \cite{li2021adaptive} introduced a Superpixel-guided Clustering (SGC) module to generate adaptive multiple prototypes. DCP \cite{lang2022beyond} generated prototypes by partitioning the support mask into four segments, using the segmentation decoder directly on the support features.

Notwithstanding, these prototype-based techniques only consider the support features during prototype generation. This approach might limit specificity due to potential variances in the appearance of target objects in support and query images. In contrast, our proposed method addresses this issue by considering both support and query features during prototype generation, thereby producing customized prototypes apt for the current query.

\section{Proposed Method}

\subsection{Problem Definition}

FSS aims to train a model on base classes and segment objects from novel classes in the query image, guided by a few annotated support samples. Specifically, the entire dataset is partitioned into two distinct subsets, $D_{base}$ and $D_{novel}$. In $D_{base}$, the foreground objects in all images originate from the base classes $C_{base}$, while in $D_{novel}$, the foreground objects in all images derive from the novel classes $C_{novel}$.
During training, $D_{base}$ is partitioned into several episodes, each containing $K+1$ randomly sampled image-mask pairs featuring the same foreground class. Among them, $K$ pairs constitute the support set $\mathcal{S}=\left\{(I^i_s,M^i_s)\right\}_{i=1}^K$, where $I^i_s\in \mathbb{R}^{H\times W\times 3}$ denotes a support image and $M^i_s\in \mathbb{R}^{H\times W}$ signifies its corresponding ground-truth mask, with $H\times W$ indicating the spatial size. The remaining sample forms the query set $Q=\left\{(I_q, M_q)\right\}$, where $I_q\in \mathbb{R}^{H\times W\times 3}$ designates the query image and $M_q\in \mathbb{R}^{H\times W}$ represents its ground-truth mask. Through this episodic training, the network develops the ability to segment a new class with only a few annotated support samples. During inference, $D_{novel}$ is also partitioned into episodes for evaluation in a similar manner, except that the ground-truth query mask $M_q$ is excluded.

\subsection{Overview}

Fig. \ref{overview} depicts the overview of the proposed QPENet, comprised of three key components: the foreground prototype evolution, the background prototype evolution, and the feature filtering and activation module for decoding. Foreground prototype evolution is achieved by the PPG and DPE modules, and the background prototype evolution is facilitated by the GBC module. Specifically, given the support image $I_s$ and the query image $I_q$, two weight-shared convolutional neural networks (CNNs) are employed to extract the support features $F_s \in \mathbb{R}^{h\times w \times c}$ and query features $F_q \in \mathbb{R}^{h\times w \times c}$, respectively, where $h\times w$ indicates the spatial size and $c$ denotes the number of channels. Then, the PPG module generates a pseudo-prototype $P^{psd}_{fg}$, which encapsulates the foreground features of the query. Next, the DPE module uses $P^{psd}_{fg}$ to reverse-segment the support image, yielding two evolved foreground prototypes, \ie, the main prototype $P^{main}_{fg}$ and the auxiliary prototype $P^{aux}_{fg}$. In parallel, the GBC module generates a customized background prototype $P_{bg}$, which adapts to the current query, based on the global background prototype $P^{global}_{bg}$, discarding any elements potentially representing the current foreground class. Finally, the prototypes and query feature $F_q$ are input into the FFA module to produce the final prediction $y^{final}$ for the query image.

\subsection{Foreground Prototype Evolution}
\label{fg_evo}

Most previous prototype-based FSS methods \cite{wang2019panet, liu2020part, yang2020prototype, tian2020prior, zhang2021self} exclusively rely on the support features to generate prototypes for feature matching and segmentation. Although these prototypes capture the information from the support images, they ignore the unique requirements of individual query image. Despite the common foreground class between the support and query images, object appearances can vary due to discrepancies in angles, positions, and the included parts of the object within images (\eg, in Fig. \ref{overview}, while the support image contains an entire airplane, the query image only includes the airplane's nose). 
Therefore, different parts of the support features hold different levels of importance for segmenting the current query image. It should be determined by the specific query image about which parts to be included in the prototype.
Guided by this understanding, we develop a strategy for query-guided prototype evolution. This section delves into the evolution of foreground prototypes, and the evolution of background prototypes will be introduced in Section \ref{bg_evo}.

Our strategy for foreground prototype evolution is outlined as follows: Initially, a prototype sourced solely from the support features is applied to perform a preliminary segmentation of the query image. This initial segmentation outcome contributes to the generation of a pseudo-prototype, which is then used to reverse-segment the support image. The reverse segmentation outcome decouple the ground-truth mask of the support image into main and auxiliary regions, thereby yielding two corresponding evolved prototypes. This iterative \textit{support-query-support} process ensures the unique requirements of the current query are encapsulated within the evolved prototypes, thereby enhancing the segmentation results. This process is realized by the mutual cooperation of two modules, \ie, the PPG module and the DPE module.

\subsubsection{Pseudo-prototype Generation Module}
Within the PPG module, the initial step involves generating a preliminary prototype based on the support features and the ground-truth support mask:
\begin{equation}
P_{fg}^{pre} = MAP(F_s, \hat{M}_s) = \frac{\sum_{i=1}^{h\times w} F_s(i) \otimes \hat{M}_s(i)}{\sum_{i=1}^{h\times w} \hat{M}_s(i)},
\end{equation}
where $MAP$ denotes a masked average pooling (MAP) operation, $i$ indicates the spatial location, $\otimes$ signifies the pixel-wise multiplication, and $\hat{M}_s \in \mathbb{R}^{h\times w}$ represents the downsampled ground-truth support mask that is spatially aligned with $F_s$. Analogous to prior works, the resulting prototype $P_{fg}^{pre}$ condenses all the foreground features of the support into a single representation. 
In our approach, we consider this prototype as a preliminary prototype, from which we derive further customized prototypes to accommodate the unique requirements of the current query.

After achieving $P_{fg}^{pre}$, we expand it to match the spatial dimensions of the query features $F_q$ and concatenate them. Subsequently, several convolutional layers are employed to make the preliminary prediction for the query mask, as described by the following equation:
\begin{equation}
y^{pre}_{fg}=Convs(Concat(F_q, BC(P_{fg}^{pre}))).
\end{equation}
Here, $BC$ signifies the broadcasting operation used to align the spatial size, $Concat$ denotes concatenation. $Convs$  consists of $3\times3$ and $1\times1$ convolutional layers.

Next, we employ $y^{pre}_{fg}$ to perform a MAP operation on the query features, yielding a pseudo-prototype that incorporates the foreground features of the query:
\begin{equation}
P_{fg}^{psd} = MAP(F_q, y^{pre}_{fg}) = \frac{\sum_{i=1}^{h\times w} F_q(i) \otimes y^{pre}_{fg}(i)}{\sum_{i=1}^{h\times w} y^{pre}_{fg}(i)}.
\end{equation}
When segmenting the query image, we primarily depend on those portions of the support features that are consistent with the query's foreground, while the remaining parts play an auxiliary role. For instance, in the example depicted in Fig. \ref{overview}, the segmentation of the query image primarily relies on the features of the airplane's nose, while the other parts carry less significance. Hence, we could interpret $P_{fg}^{psd}$ as embodying the unique requirements of the current query segmentation. In the ensuing DPE module, we will identify the pertinent parts within the support features based on the requirements reflected by this pseudo-prototype, thereby deriving evolved prototypes that can be viewed as custom solutions to these requirements.

\subsubsection{Dual Prototype Evolution Module}
At the beginning of the DPE module, the pseudo-prototype produced by the PPG module is employed to reverse-segment the foreground mask of the support image:
\begin{equation}
y^{rev}_{fg}=Convs(Concat(F_s, BC(P_{fg}^{psd}))).
\end{equation}
This reverse segmentation outcome $y^{rev}_{fg}$ emphasizes regions of the support image that correlate with the foreground regions of the query.

Utilizing $y^{rev}_{fg}$, we decouple the ground-truth support mask $M_s$ into two distinct regions. The first region encompasses the overlap with $y^{rev}_{fg}$, while the second region excludes areas covered by $y^{rev}_{fg}$. These two regions can be represented as follows:
\begin{equation} 
    M^{main}_s=\mathbbm{1} \{M_s=1\} \otimes \mathbbm{1} \{y^{rev}_{fg}=1\},
\end{equation}
\begin{equation} 
    M^{aux}_s=\mathbbm{1} \{M_s=1\} \otimes \mathbbm{1} \{y^{rev}_{fg}\neq 1\}.
\end{equation}
where $\mathbbm{1}\left \{ \cdot \right \}$ denotes a function that outputs 1 if the condition is met or 0 otherwise.
Here, $M^{main}_s$ is the common region between $M_s$ and $y^{rev}_{fg}$, reflecting the regions in the support that share similar characteristics with the query and are thus most likely to provide assistance in segmenting the query. Using $M^{main}_s$, we generate the evolved prototype $P^{main}_{fg}$:
\begin{equation}
P_{fg}^{main} = MAP(F_s, M^{main}_s) = \frac{\sum_{i=1}^{h\times w} F_s(i) \otimes M^{main}_s(i)}{\sum_{i=1}^{h\times w} M^{main}_s(i)}.
\end{equation}
Tailored to cater to the core requirements of the query, $P^{main}_{fg}$ will provide the maximum assistance in segmentation. Conversely, $M^{aux}_s$ represents regions within $M_s$ but outside $y^{rev}_{fg}$. Although these features may not contribute substantially to query segmentation, we employ $M^{aux}_s$ to generate an auxiliary prototype $P_{fg}^{aux}$ considering that the pseudo-prototype originates from a preliminary and imperfect segmentation. It can provide supplementary information to compensate for any potential information loss instigated by errors in the initial segmentation:
\begin{equation}
P_{fg}^{aux} = MAP(F_s, M^{aux}_s) = \frac{\sum_{i=1}^{h\times w} F_s(i) \otimes M^{aux}_s(i)}{\sum_{i=1}^{h\times w} M^{aux}_s(i)}.
\end{equation}

Through the joint operation of the PPG and DPE modules, the initial non-specific prototype, which was solely based on the support features, evolves into a tailored prototype addressing the unique needs of the current query. In the ensuing segmentation process, multiple foreground prototypes from different evolution stages collaborate, with $P_{fg}^{main}$ assuming a central role, while $P_{fg}^{aux}$ and $P_{fg}^{pre}$ complement by compensating in any its information omission. 

\subsection{Background Prototype Evolution}
\label{bg_evo}

\begin{figure}[t]
    \center{\includegraphics[width=\linewidth]{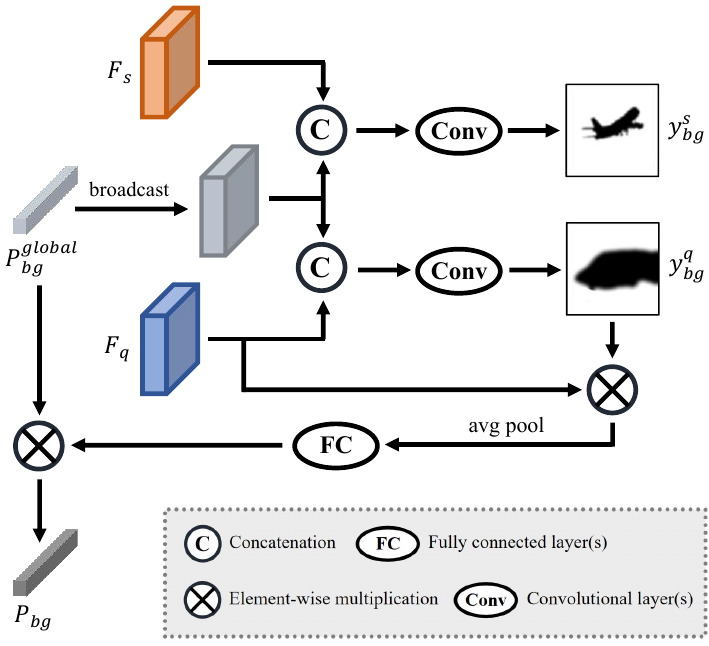}} 
    \caption{
    The detailed architecture of the GBC module. It initially uses the global background prototype to segment the backgrounds of both the support and query images. With the assistance of the predicted query background mask, it then extracts the background features of the query. Finally, these background features enable the generation of a customized background prototype tailored to the specific query after undergoing fully connected layers.}
    \label{gbc}
\end{figure}

Many existing methods \cite{lang2022beyond, yang2020prototype} generate the background prototype by directly applying MAP to the background regions of the support features. However, given that the background semantics of the query image can significantly differ from those of the support image, the produced background prototype may fail to align with the background semantics of the current query image. This discrepancy makes the accurate elimination of background components from the query features challenging. To tackle this challenge, \cite{liu2022learning} incorporated the generalized background semantics from all training images into a global background prototype, which increases the likelihood of matching the background of the current query. Nevertheless, this solution introduces a new issue: the current foreground class may appear as the background in other support images, implying that the global background prototype could contain components representing the current foreground class. If applied, such a global background prototype could wrongly suppress the foreground regions of the current query features.

To this end, we also develop a query-guided evolution strategy for the background prototype, primarily facilitated by the proposed GBC module, as depicted in Fig. \ref{gbc}. The GBC module takes the global background prototype as the input and leverages the query to eliminate potential components that resemble the current foreground. This procedure enables the generation of a background prototype more aptly attuned to the current episode. During training, the GBC module also undertakes the responsibility of updating the global background prototype by integrating the background features of both the current support and query images.

At the outset, the global background prototype is a randomly initialized vector. During training, it is supervised to predict the background map for both query and support images. Throughout this process, the vector is continuously updated to progressively integrate background information from all training images, thereby forming a robust global background prototype.
Specifically, we initially employ the global background prototype $P_{bg}^{global}$ to generate the background mask of the support image, denoted as $y^{s}_{bg}$, and the background mask of the query image, designated as $y^{q}_{bg}$. These operations are defined by the following equations:
\begin{equation}
y^{s}_{bg}=Convs(Concat(F_s, BC(P_{bg}^{global}))),
\end{equation}
\begin{equation}
y^{q}_{bg}=Convs(Concat(F_q, BC(P_{bg}^{global}))).
\end{equation}
These operations accomplish two objectives: first, they facilitate the update of the global background prototype parameters under the supervision of the ground-truth background mask; second, a preliminarily background mask of the query image is produced, analogous to the initial foreground segmentation in the PPG module. It's important to note that the segmentation of the support image is exclusively executed during training to add more training samples to update the global background prototype, which is omitted during testing.

Subsequently, leveraging the preliminary query background mask $y^{q}_{bg}$, we cleanse the global background prototype as follows:
\begin{equation}
P_{bg} = P_{bg}^{global} \otimes FC(FC(GAP(F_q \otimes y_{bg}^q))),
\end{equation}
where $FC$ denotes a fully connected layer, and $GAP$ is a global average pooling operation. This formula first extracts a typical background representation from $F_q$ using $y^{q}_{bg}$ and global average pooling. Similar to the foreground prototype, this segment can be perceived as reflecting the unique requirements for segmenting the current query. Ideally, these features should exclude any components of the current foreground class, emphasizing the semantic characteristics of the current query background. Consequently, generating a ``cleaner" based on these features to cleanse the global background prototype is suitable. To enhance the efficacy of this ``cleaner", we further refine it with two fully connected layers. Finally, the resulting background prototype $P_{bg}$ should be pure, containing only the background semantics that match the current query. Compared to the global background prototype, $P_{bg}$ can be considered as a customized background prototype that has evolved under the guidance of the query.

\subsection{Feature Filtering and Activation Module}

With the evolved foreground and background prototypes, a new challenge is how to effectively amalgamate the capabilities of each prototype to optimize the final segmentation results. To address this, we devise the FFA module.

First, we eliminate background components from the query features $F_q$ by:
\begin{equation}
    \dot{F}_q = Conv_1(Concat(F_q, BC(P_{bg}))),
\end{equation}
where $Conv_1$ represent $1\times1$ convolutional layer. 

Although the response of the background is partially suppressed, procuring precise foreground segmentation necessitates the comprehensive utilization of the information from all three foreground prototypes: $P_{fg}^{pre}$, $P_{fg}^{main}$, and $P_{fg}^{aux}$. Among these, the evolved $P_{fg}^{main}$ and $P_{fg}^{aux}$ encapsulate the specific requirements of the current query, making it vital to fully harness their capabilities. 
With these prototypes, directly integrating them with query features is a common approach, which requires broadcasting the prototypes to match the size of the query features and concatenating them along the channel dimension. Although this method works well for a single prototype, it becomes problematic when processing multiple prototypes together, resulting in too many merged feature channels and complicating accurate predictions. Moreover, prototype broadcasting produces more redundant information, especially when multiple prototypes are processed simultaneously. To address these issues, we apply the activation map \cite{lang2022beyond}, which calculates the cosine similarity between foreground prototypes and query features at each spatial location, thereby associating the relative relationship between the foreground prototype and query features in various spatial regions, and emphasizing vital areas within the entire feature set.
To this end, 
for each of these prototypes, we compute the cosine similarity with $F_q$ at every spatial location, thereby generating the activation maps, $A_{main} \in \mathbb{R}^{H \times W \times 1}$ and ${A}_{aux} \in \mathbb{R}^{H \times W \times 1}$:
\begin{equation}
    A_{main}(i)=\frac{{P_{fg}^{main}}^{\top} \cdot F_q(i)}{\Vert{P_{fg}^{main}}\Vert \cdot\Vert F_q(i)\Vert},
\end{equation}
\begin{equation}
    A_{aux}(i)=\frac{{P_{fg}^{aux}}^{\top} \cdot F_q(i)}{\Vert{P_{fg}^{aux}}\Vert \cdot\Vert F_q(i)\Vert}.
\end{equation}

These activation maps explicitly model the similarity between the prototype and $F_q$, thereby emphasizing similar regions that can subsequently aid the segmentation.
\begin{table*}[!t]
\setstretch{1}
\caption{Comparison against state-of-the-art methods on the PASCAL-$5^i$ dataset under 1-shot and 5-shot settings, assessed by class mIoU(\%) and FB-IoU(\%). `Mean’ indicates the averaged class mIoU all four folds. 
The best performance is marked in \textbf{BOLD}, and the second best performance is marked in \underline{UNDERLINE}.}
\label{pascal_comp_miou}
\resizebox{\textwidth}{44mm}{
\begin{tabular}{c|c|cccccc|cccccc}
\hline
\multirow{2}{*}{Backbone}   & \multirow{2}{*}{Methods}          & \multicolumn{6}{c|}{1-shot}                           & \multicolumn{6}{c}{5-shot}               \\
                            &                                   & Fold-0 & Fold-1 & Fold-2 & Fold-3 & Mean & FB-IoU    & Fold-0 & Fold-1 & Fold-2 & Fold-3 & Mean & FB-IoU \\ 
                            \hline
\multirow{12}{*}{ResNet-50} & CANet \cite{zhang2019canet}       & 52.5   & 65.9   & 51.3   & 51.9   & 55.4  &  66.2          & 55.5   & 67.8   & 51.9   & 53.2   & 57.1 & 69.6  \\
                            & PPNet \cite{liu2020part}          & 48.6   & 60.6   & 55.7   & 46.4   & 52.8  &  -          & 58.9   & 68.3   & 66.8   & 58.0   & 63.0 & -      \\
                            & RPMMs \cite{yang2020prototype}    & 55.2   & 66.9   & 52.6   & 50.7   & 56.3  &  -          & 56.3   & 67.3   & 54.5   & 51.0   & 57.3 & -     \\
                            & PFENet \cite{tian2020prior}       & 61.7   & 69.5   & 55.4   & 56.3   & 60.8  &  73.3          & 63.1   & 70.7   & 55.8   & 57.9   & 61.9 & 73.9      \\
                            & RePRI \cite{boudiaf2021few}       & 59.8   & 68.3   & 62.1   & 48.5   & 59.7  &  -          & 64.6   & 71.4   & 71.1   & 59.3   & 66.6 & -     \\
                            & ASGNet \cite{li2021adaptive}      & 58.8   & 67.9   & 56.8   & 53.7   & 59.3  &  69.2          & 63.7   & 70.6   & 64.2   & 57.4   & 63.9 & 74.2      \\
                            & CMN \cite{xie2021few}             & 64.3   & 70.0   & 57.4   & 59.4   & 62.8  &  72.3          & 65.8   & 70.4   & 57.6   & 60.8   & 63.7 & 72.8      \\
                            & SAGNN \cite{xie2021scale}         & 64.7   & 69.6   & 57.0   & 57.2   & 62.1  & 73.2  & 64.9   & 70.0   & 57.0   & 59.3   & 62.8  & 73.3\\
                            & CyCTR \cite{zhang2021few}         & 65.7   & 71.0   & 59.5   & 59.7   & 64.0  & -           & 69.3   & 73.5   & 63.8   & 63.5   & 67.5 & -\\
                            & DPNet \cite{mao2022learning}      & 60.7   & 69.5   & 62.8   & 58.0   & 62.7  & -           & 64.7   & 70.8   & 69.0   & 60.1   & 66.2 & -      \\
                            & DCP \cite{lang2022beyond}         & 63.8   & 70.5   & 61.2   & 55.7   & 62.8  & 75.6           & 67.2   & 73.1   & 66.4   & 64.5   & \underline{67.8} & \underline{79.7}      \\
                            & NERTNet \cite{liu2022learning}    & 65.4   & 72.3   & 59.4   & 59.8   & \underline{64.2}  & \textbf{77.0}      & 66.2   & 72.8   & 61.7   & 62.2   & 65.7 & 78.4\\
                            & Ours                              & 65.2   & 71.9   & 64.1   & 59.5   & \textbf{65.2} & \underline{76.7}    & 68.4   & 74.0   & 67.4   & 65.2   & \textbf{68.8}  & \textbf{80.0}        \\ 
                            \hline
\multirow{9}{*}{ResNet-101} & DAN \cite{wang2020few}            & 54.7   & 68.6   & 57.8   & 51.6   & 58.2  & 71.9           & 57.9   & 69.0   & 60.1   & 54.9   & 60.5 & 72.3  \\
                            & PPNet \cite{liu2020part}          & 52.7   & 62.8   & 57.4   & 47.7   & 55.2  & 70.9           & 60.3   & 70.0   & 69.4   & 60.7   & 65.1 & 77.5  \\
                            & PFENet \cite{tian2020prior}       & 60.5   & 69.4   & 54.4   & 55.9   & 60.1  & 72.9           & 62.8   & 70.4   & 54.9   & 57.6   & 61.4 & 73.5  \\
                            & RePRI \cite{boudiaf2021few}       & 59.6   & 68.6   & 62.2   & 47.2   & 59.4  & -           & 66.2   & 71.4   & 67.0   & 57.7   & 65.6 & -  \\
                            & ASGNet \cite{li2021adaptive}      & 59.8   & 67.4   & 55.6   & 54.4   & 59.3  & 71.7           & 64.6   & 71.3   & 64.2   & 57.3   & 64.4 & 75.2  \\
                            & CyCTR  \cite{zhang2021few}        & 69.3   & 72.7   & 56.5   & 58.6   & \underline{64.3}  & 72.9           & 73.5   & 73.2   & 60.1   & 66.8   & \underline{67.0} & 75.0  \\
                            & NERTNet \cite{liu2022learning}    & 65.5   & 71.8   & 59.1   & 58.3   & 63.7  & \underline{75.3}           & 67.9   & 73.2   & 60.1   & 66.8   & \underline{67.0} & \underline{78.2}  \\
                            & Ours                              & 67.0   & 73.2   & 63.7   & 60.1   & \textbf{66.0} & \textbf{77.1}   & 69.8   & 75.5   & 66.8   & 66.3   & \textbf{69.6} & \textbf{81.1}  \\ 
                            \hline
\end{tabular}}
\label{table-compare_value}
\end{table*}

Then, we apply the activation maps to $\dot{F}_q$, where the background has been suppressed, to further delineate the foreground regions:
\begin{equation}
    \ddot{F}_q = Conv_1(Concat(\dot{F}_q, BC(P_{fg}^{pre}), A_{main}, {A}_{aux})).
\end{equation}
Given that $A_{main}$ corresponds to regions in the support and query features exhibiting high similarity, it gradually achieves higher weight during training. $A_{aux}$ functions as a compensatory factor, remedying the omissions of $A_{main}$ engendered by imprecise preliminary segmentation. Similarly, the preliminary prototype $P_{fg}^{pre}$ is also integrated into this process, allowing the network to autonomously determine the necessity of obtaining some information from this pre-evolved prototype. While it may lack specificity to the current query, certain generalized components might still offer some assistance in segmentation.

So far, we have efficiently harnessed the background and foreground prototypes to derive purified foreground features. Nonetheless, as all processing relies on prototypes, despite their capability to condense features, they may inevitably overlook certain per-pixel information. To this end, we employ a pixel-wise interaction using the cross attention mechanism \cite{hou2019cross, fu2019dual, zhu2019asymmetric} with the foreground of the support features:
\begin{equation}
    y^{final} = Convs(Cross(Cross(\ddot{F}_q, F_s\otimes M_{s}), F_s\otimes M_{s})),
\end{equation}
where $Cross$ denotes the cross attention. These dual cross attention operations thoroughly contemplate the relationship between each support-query pixel pair, thereby supplementing the per-pixel features potentially omitted in the prototypes due to the MAP process, thus further refining the results.

\subsection{Training Loss}

For the foreground parts, we employ three binary cross-entropy losses to supervise the preliminary segmentation of the query image ($y_{fg}^{pre}$), the reverse-segmentation of the support image ($y_{fg}^{rev}$), and the final prediction ($y^{final}$). The formula is given as follows:
\begin{equation}
\begin{aligned}
    L_{fg}=L_{bce}(y_{fg}^{pre}, M_q) + L_{bce}(y_{fg}^{rev}, M_s) + L_{bce}(y^{final}, M_q),
\end{aligned}
\end{equation}
where $L_{bce}$ denotes the binary cross-entropy loss.

For the background parts, we incorporate the Background Mining loss \cite{liu2022learning}, denoted as $L_{bm}$, to supervise the prediction of both the support and query background masks, $y_{bg}^s$ and $y_{bg}^q$:
\begin{equation}
    L_{bg}=L_{bm}(y_{bg}^s, M_s) + L_{bm}(y_{bg}^q, M_q).
\end{equation}

Altogether, our comprehensive training loss is computed as:
\begin{equation}
L=\alpha L_{fg} + \beta L_{bg},
\end{equation}
where $\alpha$ and $\beta$ are balanced hyperparameters.

\section{Experiments}
\subsection{Dataset and Evaluation Metric}

\textbf{Datasets}. We evaluate our model on two benchmark datasets, \ie, the PASCAL-$5^i$ dataset \cite{shaban2017one} and the COCO-$20^i$ dataset \cite{nguyen2019feature}. PASCAL-$5^i$ is constructed upon the PASCAL VOC 2012 dataset \cite{everingham2010pascal} and external annotations from SDS \cite{hariharan2011semantic}, comprising a total of 20 semantic categories. COCO-$20^i$ is a more expansive dataset, founded on the MS-COCO \cite{lin2014microsoft} dataset, encompassing 80 classes. Following \cite{tian2020prior, yang2021mining, xie2021scale, liu2022dynamic}, we partition the classes evenly into 4 folds for both datasets and perform a cross-validation across all folds.

\textbf{Evaluation Metric}. In line with prior methods \cite{li2021adaptive, yang2021mining, zhang2021self}, we adopt the class mean intersection over union (mIoU) as our primary evaluation criterion for both ablation studies and comparisons. To present a more compelling validation, we further report the foreground-background IoU (FB-IoU), a metric that evaluates on the performance of target and non-target regions, regardless of category distinctions.

\begin{table*}[!t]
\setstretch{1}
\caption{Comparison against state-of-the-art methods on the COCO-$20^i$ dataset under 1-shot and 5-shot settings, assessed by class mIoU(\%) and FB-IoU(\%). `Mean’ indicates the averaged class mIoU all four folds. 
The best performance is marked in \textbf{BOLD}, and the second best performance is marked in \underline{UNDERLINE}.}
\label{coco_comp_miou}
\resizebox{\textwidth}{36mm}{

\begin{tabular}{c|c|cccccc|cccccc}
\hline
\multirow{2}{*}{Backbone}   & \multirow{2}{*}{Methods} & \multicolumn{6}{c|}{1-shot}              & \multicolumn{6}{c}{5-shot}               \\
                            &                          & Fold-0 & Fold-1 & Fold-2 & Fold-3 & Mean & FB-IoU  & Fold-0 & Fold-1 & Fold-2 & Fold-3 & Mean & FB-IoU \\ \hline
\multirow{10}{*}{ResNet-50} & PPNet \cite{liu2020part}                   & 28.1   & 30.8   & 29.7   & 27.7  & 29.0 & -       & 39.0   & 40.8   & 37.1   & 37.3   & 38.5  & -\\
                            & RPMMs \cite{yang2020prototype}            & 29.5   & 36.8   & 28.9   & 27.0  & 30.6 & -       & 33.8   & 42.0   & 33.0   & 33.3   & 35.5  & -\\
                            & ASGNet \cite{li2021adaptive}              & -      & -      & -      & -     & 34.6 & 60.4    & -      & -      & -      & -      & 42.4  & 67.0\\
                            & RePRI \cite{boudiaf2021few}               & 31.2   & 38.1   & 33.3   & 33.0  & 34.0 & -  & 38.5   & 46.2   & 40.0   & 43.6   & 42.1  & -\\
                            & CyCTR \cite{zhang2021few}                 & 38.9   & 43.0   & 39.6   & 39.8  & 40.3 & -  & 41.1   & 48.9   & 45.2   & 47.0   & 45.6  & -\\
                            & CMN \cite{xie2021few}                   & 37.9   & 44.8   & 38.7   & 35.6  & 39.3 & 61.7  & 42.0   & 50.5   & 41.0   & 38.9   & 43.1  & 63.3\\
                            & DPNet \cite{mao2022learning}              & -      & -      & -      & -     & 37.2 & -  & -      & -      & -      & -      & 42.9  & -\\
                            & DCP \cite{lang2022beyond}                 & 40.9   & 43.8   & 42.6   & 38.3  & \underline{41.4} & -  & 45.8   & 49.7   & 43.7   & 46.6   & \underline{46.5}  & -\\
                            & NERTNet \cite{liu2022learning}            & 36.8   & 42.6   & 39.9   & 37.9  & 39.3 & \textbf{68.5}  & 38.2   & 44.1   & 40.4   & 38.4   & 40.3  & \underline{69.2} \\
                            & Ours                                      & 41.5   & 47.3   & 40.9   & 39.4  & \textbf{42.3} & \underline{67.4}  & 47.3   & 52.4   & 44.3   & 44.9   & \textbf{47.2}   & \textbf{69.5}   \\ \hline
\multirow{6}{*}{ResNet-101} & DAN \cite{wang2020few}                    & -      & -      & -      & -     & 24.4 & 62.3  & -      & -      & -      & -      & 29.6  & 63.9\\
                            & PFENet \cite{tian2020prior}               & 34.3   & 33.0   & 32.3   & 30.1  & 32.4 & 58.6  & 38.5   & 38.6   & 38.2   & 34.3   & 37.4  & 61.9\\
                            & SCL \cite{zhang2021self}                  & 36.4   & 38.6   & 37.5   & 35.4  & 37.0 & -  & 38.4   & 40.5   & 41.5   & 38.7   & 39.9  & -\\
                            & SAGNN \cite{xie2021scale}                & 36.1   & 41.0   & 38.2   & 33.5  & 37.2 & 60.9  & 40.9   & 48.3   & 42.6   & 38.9   & 42.7  & 63.4\\
                            & NERTNet \cite{liu2022learning}            & 38.3   & 40.4   & 39.5   & 38.1  & \underline{39.1} & \underline{67.5}  & 42.3   & 44.4   & 44.2   & 41.7   & \underline{43.2}  & \underline{69.6} \\
                            & Ours                                      & 39.8   & 45.4   & 40.5   & 40.0   & \textbf{41.4}  & \textbf{67.8}  & 47.2   & 54.9   & 43.4   & 45.4       & \textbf{47.7}      & \textbf{70.6} \\ \hline
\end{tabular}
}
\label{table-compare_value}
\end{table*}

\subsection{Implementation Details}

The implementation of our code is performed using PyTorch \cite{paszke2019pytorch}, with all experiments accelerated on an NVIDIA RTX 3090 GPU. In line with previous works \cite{fan2022self, zhang2021few}, we evaluate our network using both ResNet-50 and ResNet-101 \cite{he2016deep} as backbones. Both backbones, pre-trained on ImageNet \cite{russakovsky2015imagenet}, maintain their fixed weights during the training process. During both training and inference stages, the spatial size of input images is $473 \times 473$. We train the model for 200 epochs on PASCAL-$5^i$ and 50 epochs on COCO-$20^i$, with SGD serving as the optimizer and a batch size of 4. The initial learning rate, weight decay, and momentum are respectively set at 0.0025, 0.0001, and 0.9. Meanwhile, a polynomial annealing policy with the power set to 0.9 is applied for learning rate decay. During the testing stage, we conform to the protocol established by \cite{tian2020prior}, randomly sampling 1,000 support-query pairs on PASCAL-$5^i$ or 4,000 pairs on COCO-$20^i$ from the testing fold for evaluation.

\subsection{Comparison with State-of-the-art Methods}

We compare the proposed QPENet with previous state-of-the-art FSS methods, including CANet \cite{zhang2019canet}, PPNet \cite{liu2020part}, RPMMs \cite{yang2020prototype}, PFENet \cite{tian2020prior}, RePRI \cite{boudiaf2021few}, ASGNet \cite{li2021adaptive}, CMN \cite{xie2021few}, SCL \cite{zhang2021self}, CyCTR \cite{zhang2021few}, DPNet \cite{mao2022learning}, DCP \cite{lang2022beyond}, NERTNet \cite{liu2022learning}, DAN \cite{wang2020few} and SAGNN \cite{xie2021scale}. 

\subsubsection{Quantitative Analysis}
Both mIoU and FB-IoU are considered for quantitative evaluation, the results are reported in Table \ref{pascal_comp_miou} and Table \ref{coco_comp_miou}.

\textbf{PASCAL-$5^i$ Dataset}. 
In Table \ref{pascal_comp_miou}, we compare the class mIoU and FB-IoU of our QPENet with prior methods under 1-shot and 5-shot settings on the PASCAL-$5^i$ dataset, utilizing ResNet-50 and ResNet-101 as backbones. Overall, our model demonstrates satisfactory performance. With ResNet-50 as the backbone, our QPENet surpasses the previous best record by 1.0\% class mIoU under the 1-shot setting and 1.0\% class mIoU under the 5-shot setting. When employing the ResNet-101 backbone, our model also performs best under both 1-shot and 5-shot settings, delivering a class mIoU enhancement of 1.7\% and 2.6\% respectively, relative to the preceding top score. Moreover, when assessed by FB-IoU, our model consistently excels, achieving the highest ranking in three out of the four combinations of backbones and settings, and securing the second place in the remaining one. The consistent high-performance of our model across diverse datasets, settings, and backbones attests to its robustness and reliability, validating the correctiveness of our ideas.

\textbf{COCO-$20^i$ Dataset}. 
The COCO-$20^i$ benchmark presents a more demanding challenge, with each image typically containing multiple objects and exhibiting a more diverse and complex set of scenes. The performance of various methods on this dataset is summarized in Table \ref{coco_comp_miou}. As indicated by these results, our model excels in both 1-shot and 5-shot settings. Specifically, in the 1-shot setting, our method yields mean class mIoU scores of 42.3\% and 41.4\% using the two backbones, surpassing the prior top results by respective margins of 0.9\% and 2.3\%. Under the 5-shot setting, our method outperforms the state-of-the-art by 0.7\% and 4.5\% with the two respective backbones. 
Regarding FB-IoU, akin to the situation on PASCAL-$5^i$, our QPENet secures the first place in three of the four combinations of backbones and settings, and attains the second position in the remaining one. 
The strong performance on the COCO-$20^i$ dataset emphasizes the ability of our model to handle demanding situations and produce commendable results.

\begin{figure*}[htbp]
	\center{\includegraphics[width=18cm]  {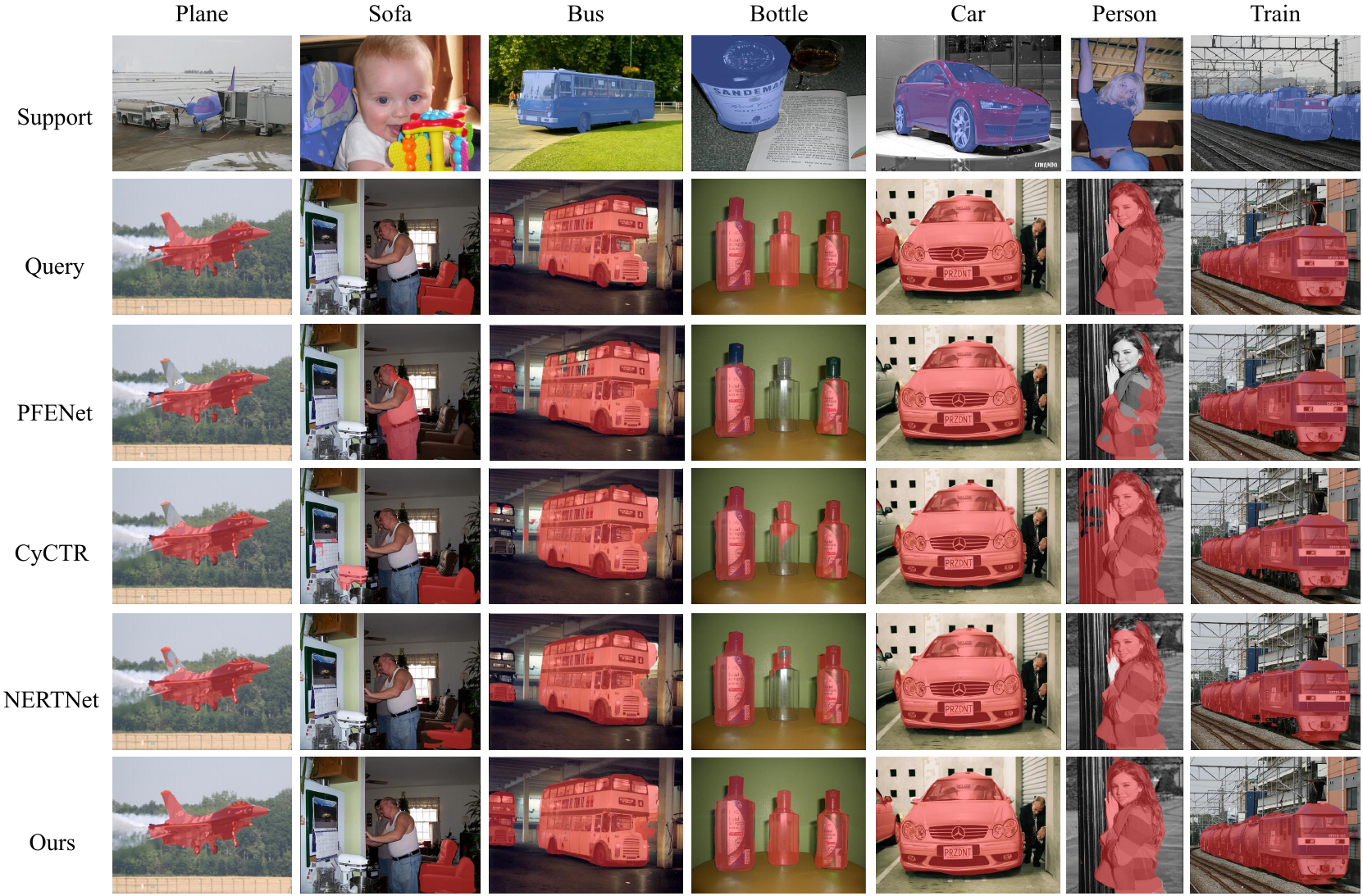}} 
	\caption{Qualitative comparison against state-of-the-art methods in various representative scenes. From top to bottom: annotated support image; annotated query image; predictions of PFENet; predictions of CyCTR; predictions of NERTNet; predictions of our QPENet.}
	\label{qual_comp}
\end{figure*}

\subsubsection{Qualitative Analysis}
In Fig. \ref{qual_comp}, we illustrate a selection of visualized results produced by our QPENet and prior FSS models \cite{tian2020prior,zhang2021few,liu2022learning} on both PASCAL-$5^i$ and COCO-$20^i$ benchmarks. For a more convincing comparison, we have chosen seven distinct scenes to exemplify the overall performance of each model. These examples highlight the superior performance of our method in terms of both prediction completeness and accuracy. In scenarios where multiple instances from the target category are present in the query image, competing methods often fail to identify less conspicuous ones. Contrarily, our model effectively recognizes all instances of the target category, as exemplified by the \textit{Bus} and \textit{Car} scenarios. This can be attributed to the fact that the prototypes in other models are generalized representation that evenly incorporate the support features. For an instance wherein only a small section of it appears in the image, the features can significantly deviate from the generalized representation of the category. For instance, in the \textit{Bus} scenario, the smaller bus is displayed solely by its frontal view, while in the \textit{Car} example, the overlooked car instance presents only a side view. Hence, the specific features of these instances differ markedly from the typical characteristics provided by the prototypes, leading to potential omissions. In contrast, our model considers the characteristics of all query instances to be segmented and matches the corresponding solutions from the support features, thus avoiding any omissions. For images containing a single instance of the target category, the superior performance of our model is also evident. As demonstrated by the \textit{Person} and \textit{Train} examples, the predictions of our QPENet tend to be more precise and complete. This is due to the more targeted prototype formulation, and the optimized prototype utilization of the FFA module, yielding a more precise performance compared to models that resort to an average representation of all foreground regions within the support features.

\begin{table}[!t]
\centering
\renewcommand\arraystretch{1.2}
    \begin{spacing}{1}
    \caption{Ablation study results under different module configurations, assessed by class mIoU(\%) and FB-IoU(\%).}
    \label{abla_quan}
    \setlength{\tabcolsep}{0.3mm}{
        \begin{tabular}{p{1.18cm}<{\centering}p{1.18cm}<{\centering}p{1cm}<{\centering}|p{0.96cm}<{\centering}p{1.18cm}<{\centering}p{0.96cm}<{\centering}p{0.96cm}<{\centering}p{0.96cm}<{\centering}}
        \hline
        \textit{FGPE}   & \textit{BGPE} & \textit{FFA}  & mIoU & FB-IoU\\ 
        \hline
                        &               &               & 62.4 & 74.1 \\
        \checkmark      &               &               & 64.1 & 75.7 \\
        \checkmark      & \checkmark    &               & 64.5 & 76.2 \\
        \checkmark      &               & \checkmark    & 64.8 & 76.3 \\
        \checkmark      & \checkmark    & \checkmark    & 65.2 & 76.7 \\

        \hline
        \end{tabular}}
    \end{spacing}
\end{table}

\subsection{Ablation Study}
In this section, we report ablation study results of our QPENet. Following \cite{liu2022intermediate, liu2022dynamic}, these experiments are performed under the 1-shot setting of PASCAL-$5^i$, using the ResNet-50 backbone.

\begin{figure*}[htbp]
	\center{\includegraphics[width=18cm]  {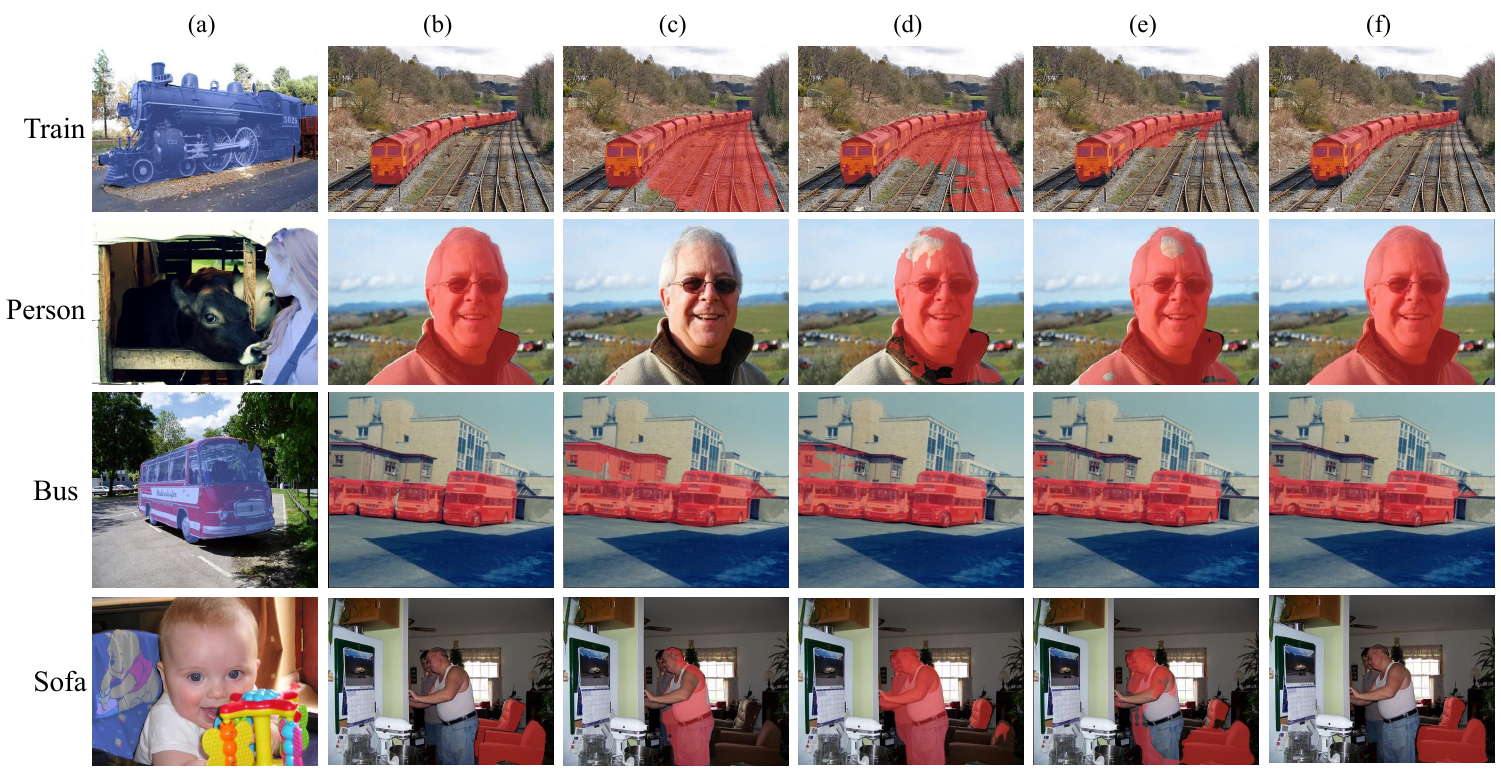}} 
	\caption{Qualitative results for component analysis. (a) Annotated support image. (b) Annotated query image. (c) Predictions of the baseline model. (d) Predictions of the baseline model enhanced by \textit{FGPE}. (e) Predictions of the baseline model enhanced by \textit{FGPE} and \textit{BGPE}. (f) Predictions of the full model.}
	\label{component_analysis}
\end{figure*}

\begin{figure}[t]
    \center{\includegraphics[width=\linewidth]{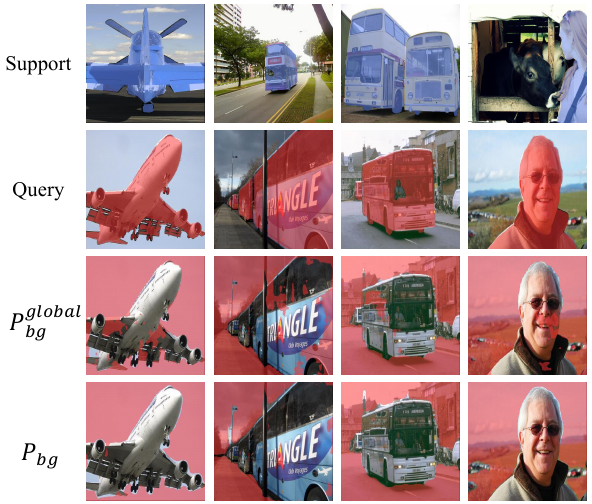}} 
    \caption{
    Visualized background segmentation results for different background prototypes. From top to bottom: annotated support image; annotated query image; segmentation results with the global background prototype $P_{bg}^{global}$; segmentation results with the evolved background prototype $P_{bg}$.
    }
    \label{bg_activate}
\end{figure}

\textbf{Effectiveness of Different Components.} 
Table \ref{abla_quan} illustrates the verification of the key components incorporated in our model, \ie, the Foreground Prototype Evolution strategy (\textit{FGPE}), the Background Prototype Evolution strategy (\textit{BGPE}), and the decoder of the Feature Filtering and Activation module (\textit{FFA}). The results in the first row derive from our baseline model, which omits both prototype evolution strategies, utilizing the preliminary foreground prototype $P_{fg}^{pre}$ and the global background prototype $P_{bg}^{global}$ \cite{liu2022learning}. Its decoder is FEM \cite{tian2020prior}, which is commonly used in the FSS field, instead of the proposed FFA module. Apart from the quantitative results in Table \ref{abla_quan}, we also offer several visual examples in Fig. \ref{component_analysis} for further elucidation.

Firstly, we enhance the baseline model with \textit{FGPE}, which encompasses our proposed PPG and DPE modules. As demonstrated in the second row of Table \ref{abla_quan}, the \textit{FGPE} yields a class mIoU increment of 1.7\%, thus validating its efficacy. By adapting the foreground prototype evolution using the query, the prototype becomes more suited for the current query segmentation task. From the visual results in Fig. \ref{component_analysis}, it can also be observed that the model with \textit{FGPE} performs better in terms of both precision and recall compared to the model without it.

Subsequently, we introduce \textit{BGPE}, primarily composed of the GBC module. The performance resulting from this modification, as depicted in the third row of Table \ref{abla_quan}, sees a further class mIoU enhancement of 0.4\%, confirming the effectiveness of guiding the background prototype evolution with query features. This strategy filters out the detrimental components potentially representing the current foreground class from the global background prototype, thereby producing a more adaptive background prototype for the current query and leading to superior performance. 
From the fifth column of Fig. \ref{component_analysis}, it can also be observed that the qualitative results improves with the incorporation of \textit{BGPE}. A typical example is the \textit{Person} scenario where, in the absence of \textit{BGPE}, the man's body is erroneously considered as background. However, the inclusion of \textit{BGPE} rectifies this issue, ensuring the complete segmentation result.
In Fig. \ref{bg_activate}, we further visualize some instances comparing the segmentation of the background using the global background prototype $P_{bg}^{global}$ and our evolved background prototype $P_{bg}$. It is apparent that $P_{bg}^{global}$ erroneously classifies some foreground regions as background, while $P_{bg}$ does not have this issue. This further demonstrates the necessity of \textit{BGPE}. 
We also compare the performance and network parameter count when implementing prototype evolution versus not implementing it. When we introduce the prototype evolution, the total number of parameters increases from 32.7 MB to 34.6 MB, while the performance index mIoU increases from 62.4 to 64.8, with an improvement of 2.4\%. Overall, this operation is cost-effective. Thus, when considering both performance enhancement and parameter count, prototype evolution proves to be an efficient method.

Finally, as displayed in the final two rows of Table \ref{abla_quan}, we replace the decoder in the preceding two configurations with \textit{FFA}, resulting in class mIoU improvements of 0.7\%. This clearly illustrates the superiority of our FFA module. The FFA module, which considers the background prototype and multiple foreground prototypes comprehensively, selectively applies them, and examines the per-pixel relationship between support and query features, indeed delivers enhanced outcomes. Additionally, the visualizations in Fig. \ref{component_analysis} also show that the full model incorporating \textit{FFA} exhibits the best performance across all four scenarios.

\textbf{Influence of Different Foreground Prototypes.} 
Table \ref{as-1} shows the influence of different foreground prototypes within the proposed QPENet. 
If only $P_{fg}^{pre}$ is adopted, the mIoU and FB-IoU socres are 63.3\% and 74.6\% respectively. When using the $P_{fg}^{main}$ and $P_{fg}^{aux}$ separately, their performance is not as good as using $P_{fg}^{pre}$ alone, with achieving 61.8\% and 58.1\% on mIoU. This is because they only represent part of the semantic information of the target category in the support image, and the semantic information is incomplete when these two prototypes are used alone. Furthermore, we also evaluate the performance when using both of $P_{fg}^{main}$ and $P_{fg}^{aux}$. We can see that the performance improves compared to either of them. This is because their semantic information is complementary. Finally, when using all support prototypes together, we achieve 65.2\% and 76.7\% on mIoU and FB-IoU, with a significant gain of 1.9\% on mIoU compared to only use $P_{fg}^{pre}$. Because $P_{fg}^{main}$ and $P_{fg}^{aux}$ highlight the parts that are identical and the parts that are different between the foreground features and the query features respectively, which makes more prior knowledge about the process of segmenting the object category in the query image. 

\begin{table}[!t]
\centering
\renewcommand\arraystretch{1.0}
    \begin{spacing}{1}
    \caption{Comparison of class mIoU(\%) and FB-IoU(\%) across different foreground prototypes.}
    \label{as-1}
    \setlength{\tabcolsep}{2mm}{
        \begin{tabular}{ccc|cc}
            \hline
            $P_{fg}^{pre}$    & $P_{fg}^{main}$    & $P_{fg}^{aux}$     & mIoU    & FB-IoU \\ \hline
            \checkmark  &                    &                    & 63.3        & 74.6          \\
                        & \checkmark         &                    & 61.8        & 72.0           \\
                        &                    & \checkmark         & 58.1        & 69.4           \\
                        & \checkmark         & \checkmark         & 64.1        & 74.7           \\
            \checkmark  & \checkmark         & \checkmark         & 65.2        & 76.7           \\ \hline
            \end{tabular}}
    \end{spacing}
\end{table}

\textbf{Influence of Activation Maps.} In Table \ref{as-2}, we present a comparative analysis of the outcomes when using activation maps against not using them. In the latter case, we directly concatenate the broadcasted prototypes with the query features, denoted as ‘w/o activation maps’. The activation maps focus more on the relative relationships between foreground prototypes and query features across different spatial areas, thereby emphasizing key regions within the entire feature. The results demonstrate that the performance with activation maps is obviously better, with mIoU and FB-IoU being higher by 3.6\% and 3.7\%, respectively, confirming the effectiveness of this strategy.

\begin{table}[!t]
\centering
\renewcommand\arraystretch{1.0}
    \begin{spacing}{1}
    \caption{Ablation study of activation maps and DPE, assessed by class mIoU (\%) and FB-IoU (\%).}
    \label{as-2}
    \setlength{\tabcolsep}{2mm}{
        \begin{tabular}{c|cc}
            \hline
             & mIoU & FB-IoU \\ \hline
            w/o activation maps           & 61.6 & 73.0   \\ \hline
            w SGM           & 63.5 & 75.6   \\ \hline
            Full Model             & 65.2 & 76.7   \\ \hline  
            \end{tabular}}
    \end{spacing}
\end{table}

\textbf{Ablation Study of DPE.} 
In our method, we design the DPE module to produce a match between the query's specific requirements and the support features.
It seems to be somewhat similar in structure to the SGM module in SCL \cite{zhang2021self}, but they have a key difference in the source of guidance information.
In SGM, the guiding information is derived from prototypes formed from the foreground portion of the support features. In contrast, our DPE employs pseudo-prototypes generated from the foreground portion of the query image, informed by preliminary predictions. Furthermore, the design objectives of DPE and SGM also differ. According to \cite{zhang2021self}, the purpose of SGM is to extract comprehensive support information from the support set. In contrast, the purpose of our DPE is to utilize a pseudo-prototype generated from query features to isolate the supporting foreground features, breaking them down into two parts with different affinities to the query features. In Table \ref{as-2}, we show the results of replacing DPE in the model with SGM, denote as ‘w SGM’. The results show that the effect of using DPE is better, the mIoU and FB-IoU are improved by 1.7\% and 1.1\%, respectively, which proves the necessity of considering query features in the process of generating foreground prototypes.

\section{Conclusion}
In this paper, we propose a novel prototype-based FSS method, named QPENet. The fundamental principle underlying QPENet is to optimize prototypes to address the specific characteristics of the current query image, thereby improving the accuracy of its segmentation. QPENet presents two types of prototypes, namely, foreground and background prototypes, both of which evolve under the guidance of the query features. Specifically, we design the PPG and DPE modules to facilitate the customization of the foreground prototype in a \textit{support-query-support} manner. For the background prototype, we design the GBC module, which filters out interfering components in the global background prototype, making it adaptive to the current query. Furthermore, to maximize the utilization of these prototypes, we design the FFA module. This module considers both the foreground and background prototypes comprehensively and models the per-pixel relationship between support and query features, thereby outputting superior segmentation results. 
In addition, extensive experiments on PASCAL-$5^i$ and COCO-$20^i$ datasets demonstrated the superiority of our proposed model.


%





\ifCLASSOPTIONcaptionsoff
  \newpage
\fi



%
\bibliographystyle{IEEEtran}
\bibliography{ref}

\begin{thebibliography}{10}
\providecommand{\url}[1]{#1}
\csname url@samestyle\endcsname
\providecommand{\newblock}{\relax}
\providecommand{\bibinfo}[2]{#2}
\providecommand{\BIBentrySTDinterwordspacing}{\spaceskip=0pt\relax}
\providecommand{\BIBentryALTinterwordstretchfactor}{4}
\providecommand{\BIBentryALTinterwordspacing}{\spaceskip=\fontdimen2\font plus
\BIBentryALTinterwordstretchfactor\fontdimen3\font minus \fontdimen4\font\relax}
\providecommand{\BIBforeignlanguage}[2]{{%
\expandafter\ifx\csname l@#1\endcsname\relax
\typeout{** WARNING: IEEEtran.bst: No hyphenation pattern has been}%
\typeout{** loaded for the language `#1'. Using the pattern for}%
\typeout{** the default language instead.}%
\else
\language=\csname l@#1\endcsname
\fi
#2}}
\providecommand{\BIBdecl}{\relax}
\BIBdecl

\bibitem{tian2020prior}
Z.~Tian, H.~Zhao, M.~Shu, Z.~Yang, R.~Li, and J.~Jia, ``Prior guided feature enrichment network for few-shot segmentation,'' \emph{{IEEE} Trans. Pattern Anal. Mach. Intell.}, vol.~44, no.~2, pp. 1050--1065, 2022.

\bibitem{DBLP:journals/jei/LiGWCZW16}
C.~Li, J.~Guo, B.~Wang, R.~Cong, Y.~Zhang, and J.~Wang, ``Single underwater image enhancement based on color cast removal and visibility restoration,'' \emph{J. Electronic Imaging}, vol.~25, no.~3, p. 033012, 2016.

\bibitem{DBLP:journals/access/NiLCZPF17}
M.~Ni, J.~Lei, R.~Cong, K.~Zheng, B.~Peng, and X.~Fan, ``Color-guided depth map super resolution using convolutional neural network,'' \emph{{IEEE} Access}, vol.~5, pp. 26\,666--26\,672, 2017.

\bibitem{crm/tip20/MCMT-GAN}
Y.~Huang, F.~Zheng, R.~Cong, W.~Huang, M.~R. Scott, and L.~Shao, ``{MCMT-GAN:} multi-task coherent modality transferable {GAN} for {3D} brain image synthesis,'' \emph{{IEEE} Trans. Image Process.}, vol.~29, pp. 8187--8198, 2020.

\bibitem{DBLP:journals/tip/LiGRCHKT20}
C.~Li, C.~Guo, W.~Ren, R.~Cong, J.~Hou, S.~Kwong, and D.~Tao, ``An underwater image enhancement benchmark dataset and beyond,'' \emph{{IEEE} Trans. Image Process.}, vol.~29, pp. 4376--4389, 2020.

\bibitem{crm/jbhi22/polyp}
G.~Yue, W.~Han, B.~Jiang, T.~Zhou, R.~Cong, and T.~Wang, ``Boundary constraint network with cross layer feature integration for polyp segmentation,'' \emph{IEEE J. Biomed. Health Inform.}, vol.~26, no.~8, pp. 4090--4099, 2022.

\bibitem{chen2023saving}
J.~Chen, R.~Cong, Y.~LUO, H.~Ip, and S.~Kwong, ``Saving 100x storage: Prototype replay for reconstructing training sample distribution in class-incremental semantic segmentation,'' in \emph{Advances in Neural Information Processing Systems}, 2023.

\bibitem{crmbridgenet}
Q.~Tang, R.~Cong, R.~Sheng, L.~He, D.~Zhang, Y.~Zhao, and S.~Kwong, ``Bridgenet: {A} joint learning network of depth map super-resolution and monocular depth estimation,'' in \emph{Proc. ACM MM}, 2021, pp. 2148--2157.

\bibitem{crm/acmmm23/FPNet}
R.~Cong, M.~Sun, S.~Zhang, X.~Zhou, W.~Zhang, and Y.~Zhao, ``Frequency perception network for camouflaged object detection,'' in \emph{Proc. ACM MM}, 2021, pp. 1179--1189.

\bibitem{crm/acmmm23/SDDNet}
R.~Cong, Y.~Guan, J.~Chen, W.~Zhang, Y.~Zhao, and S.~Kwong, ``{SDDNet}: Style-guided dual-layer disentanglement network for shadow detection,'' in \emph{Proc. ACM MM}, 2023, pp. 1202--1211.

\bibitem{zhang2023controlvideo}
Y.~Zhang, Y.~Wei, D.~Jiang, X.~Zhang, W.~Zuo, and Q.~Tian, ``Controlvideo: Training-free controllable text-to-video generation,'' \emph{arXiv preprint arXiv:2305.13077}, 2023.

\bibitem{crm/tomm22/hippocampus}
H.~Huang, R.~Cong, L.~Yang, L.~Du, C.~Wang, and S.~Kwong, ``Feedback chain network for hippocampus segmentation,'' \emph{ACM Trans. Multimedia Comput. Commun. Appl.}, vol.~19, no.~3s, p. article 133, 2022.

\bibitem{wang2019learning}
Q.~Wang, C.~Yuan, and Y.~Liu, ``Learning deep conditional neural network for image segmentation,'' \emph{{IEEE} Trans. Multim.}, vol.~21, no.~7, pp. 1839--1852, 2019.

\bibitem{kang2018depth}
B.~Kang, Y.~Lee, and T.~Q. Nguyen, ``Depth-adaptive deep neural network for semantic segmentation,'' \emph{{IEEE} Trans. Multim.}, vol.~20, no.~9, pp. 2478--2490, 2018.

\bibitem{zhan2019unmanned}
C.~Zhan, H.~Hu, Z.~Wang, R.~Fan, and D.~Niyato, ``Unmanned aircraft system aided adaptive video streaming: A joint optimization approach,'' \emph{{IEEE} Trans. Multim.}, vol.~22, no.~3, pp. 795--807, 2019.

\bibitem{yao2021non}
Y.~Yao, T.~Chen, G.~Xie, C.~Zhang, F.~Shen, Q.~Wu, Z.~Tang, and J.~Zhang, ``Non-salient region object mining for weakly supervised semantic segmentation,'' in \emph{Proceedings of the IEEE Conference on Computer Vision and Pattern Recognition}, 2021, pp. 2623--2632.

\bibitem{crm/tmm22/TNet}
R.~Cong, K.~Zhang, C.~Zhang, F.~Zheng, Y.~Zhao, Q.~Huang, and S.~Kwong, ``Does {Thermal} really always matter for {RGB-T} salient object detection?'' \emph{IEEE Trans. Multimedia}, early access, doi: 10.1109/TMM.2022.3216476.

\bibitem{chen2022saliency}
T.~Chen, Y.~Yao, L.~Zhang, Q.~Wang, G.~Xie, and F.~Shen, ``Saliency guided inter-and intra-class relation constraints for weakly supervised semantic segmentation,'' \emph{{IEEE} Trans. Multim.}, vol.~25, pp. 1727--1737, 2023.

\bibitem{asgari2021deep}
S.~A. Taghanaki, K.~Abhishek, J.~P. Cohen, J.~Cohen{-}Adad, and G.~Hamarneh, ``Deep semantic segmentation of natural and medical images: A review,'' \emph{Artificial Intelligence Review}, vol.~54, no.~1, pp. 137--178, 2021.

\bibitem{krizhevsky2017imagenet}
A.~Krizhevsky, I.~Sutskever, and G.~E. Hinton, ``Imagenet classification with deep convolutional neural networks,'' \emph{Commun. {ACM}}, vol.~60, no.~6, pp. 84--90, 2017.

\bibitem{crm/tce22/covid}
R.~Cong, Y.~Zhang, N.~Yang, H.~Li, X.~Zhang, R.~Li, Z.~Chen, Y.~Zhao, and S.~Kwong, ``Boundary guided semantic larning for real-time {COVID-19} lung infection segmentation system,'' \emph{IEEE Trans. Consum. Electron.}, vol.~68, no.~4, pp. 376--386, 2022.

\bibitem{DBLP:journals/caaitrit/ZhaoZGY23}
G.~Zhao, Y.~Zhang, M.~Ge, and M.~Yu, ``Bilateral u-net semantic segmentation with spatial attention mechanism,'' \emph{{CAAI} Trans. Intell. Technol.}, vol.~8, no.~2, pp. 297--307, 2023.

\bibitem{crm/tnnls22/360SOD}
R.~Cong, K.~Huang, J.~Lei, Y.~Zhao, Q.~Huang, and S.~Kwong, ``Multi-projection fusion and refinement network for salient object detection in 360 $^{\circ }$ omnidirectional image,'' \emph{IEEE Trans. Neural Netw. Learn. Syst.}, pp. 1--13, 2023.

\bibitem{DBLP:journals/tmm/YinTYXW22}
C.~Yin, J.~Tang, T.~Yuan, Z.~Xu, and Y.~Wang, ``Bridging the gap between semantic segmentation and instance segmentation,'' \emph{{IEEE} Trans. Multim.}, vol.~24, pp. 4183--4196, 2022.

\bibitem{crm/tcyb23/instanceSOD}
J.~Chen, R.~Cong, H.~H.~S. Ip, and S.~Kwong, ``Kepsalinst: Using peripheral points to delineate salient instances,'' \emph{IEEE Trans. Cybern.}, pp. 1--14, 2023.

\bibitem{crm/ACMMM20/DMVOS}
P.~Wen, R.~Yang, Q.~Xu, C.~Qian, Q.~Huang, R.~Cong, and J.~Si, ``{DMVOS}: Discriminative matching for real-time video object segmentation,'' in \emph{Proc. ACM MM}, 2020, pp. 2048--2056.

\bibitem{DBLP:journals/tmm/LiuGMLW21}
H.~Liu, Y.~Guo, Y.~Ma, Y.~Lei, and G.~Wen, ``Semantic context encoding for accurate {3D} point cloud segmentation,'' \emph{{IEEE} Trans. Multim.}, vol.~23, pp. 2045--2055, 2021.

\bibitem{crm/tim22/covid}
R.~Cong, H.~Yang, Q.~Jiang, W.~Gao, H.~Li, C.~Wang, Y.~Zhao, and S.~Kwong, ``{BCS-Net}: Boundary, context, and semantic for automatic {COVID-19} lung infection segmentation from {CT} images,'' \emph{{IEEE} Trans. Instrum. Meas.}, vol.~71, pp. 1--11, 2022.

\bibitem{DBLP:journals/tmm/WangJSCJ16}
T.~Wang, Z.~Ji, Q.~Sun, Q.~Chen, and X.~Jing, ``Interactive multilabel image segmentation via robust multilayer graph constraints,'' \emph{{IEEE} Trans. Multim.}, vol.~18, no.~12, pp. 2358--2371, 2016.

\bibitem{crm/tcyb22/glnet}
R.~Cong, N.~Yang, C.~Li, H.~Fu, Y.~Zhao, Q.~Huang, and S.~Kwong, ``Global-and-local collaborative learning for co-salient object detection,'' \emph{IEEE Trans. Cybern.}, vol.~53, no.~3, pp. 1920--1931, 2023.

\bibitem{DBLP:journals/tmm/ZhouGLF21}
L.~Zhou, C.~Gong, Z.~Liu, and K.~Fu, ``{SAL}: Selection and attention losses for weakly supervised semantic segmentation,'' \emph{{IEEE} Trans. Multim.}, vol.~23, pp. 1035--1048, 2021.

\bibitem{DBLP:journals/tmm/ZhangLCSSK19}
T.~Zhang, G.~Lin, J.~Cai, T.~Shen, C.~Shen, and A.~C. Kot, ``Decoupled spatial neural attention for weakly supervised semantic segmentation,'' \emph{{IEEE} Trans. Multim.}, vol.~21, no.~11, pp. 2930--2941, 2019.

\bibitem{crm/ACMMM23/PICRNet}
R.~Cong, H.~Liu, C.~Zhang, W.~Zhang, F.~Zheng, R.~Song, and S.~Kwong, ``Point-aware interaction and {CNN}-induced refinement network for {RGB-D} salient object detection,'' in \emph{Proc. ACM MM}, 2023, pp. 406--416.

\bibitem{DBLP:journals/caaitrit/ZhengZZLL22}
X.~Zheng, Y.~Zhang, Y.~Zheng, F.~Luo, and X.~Lu, ``Abnormal event detection by a weakly supervised temporal attention network,'' \emph{{CAAI} Trans. Intell. Technol.}, vol.~7, no.~3, pp. 419--431, 2022.

\bibitem{crm/tcsvt22/weaklySOD}
R.~Cong, Q.~Qin, C.~Zhang, Q.~Jiang, S.~Wang, Y.~Zhao, and S.~Kwong, ``A weakly supervised learning framework for salient object detection via hybrid labels,'' \emph{IEEE Trans. Circuits Syst. Video Technol.}, vol.~33, no.~2, pp. 534--548, 2023.

\bibitem{vinyals2016matching}
O.~Vinyals, C.~Blundell, T.~Lillicrap, K.~Kavukcuoglu, and D.~Wierstra, ``Matching networks for one shot learning,'' in \emph{Advances in Neural Information Processing Systems}, 2016, pp. 3630--3638.

\bibitem{zhang2020sg}
X.~Zhang, Y.~Wei, Y.~Yang, and T.~S. Huang, ``{SG-One}: Similarity guidance network for one-shot semantic segmentation,'' \emph{{IEEE} Trans. Cybern.}, vol.~50, no.~9, pp. 3855--3865, 2020.

\bibitem{yang2020prototype}
B.~Yang, C.~Liu, B.~Li, J.~Jiao, and Q.~Ye, ``Prototype mixture models for few-shot semantic segmentation,'' in \emph{Proceedings of the European Conference on Computer Vision}, vol. 12353, 2020, pp. 763--778.

\bibitem{liu2020part}
Y.~Liu, X.~Zhang, S.~Zhang, and X.~He, ``Part-aware prototype network for few-shot semantic segmentation,'' in \emph{Proceedings of the European Conference on Computer Vision}, vol. 12354, 2020, pp. 142--158.

\bibitem{li2021adaptive}
G.~Li, V.~Jampani, L.~Sevilla{-}Lara, D.~Sun, J.~Kim, and J.~Kim, ``Adaptive prototype learning and allocation for few-shot segmentation,'' in \emph{Proceedings of the IEEE Conference on Computer Vision and Pattern Recognition}, 2021, pp. 8334--8343.

\bibitem{zhang2021self}
B.~Zhang, J.~Xiao, and T.~Qin, ``Self-guided and cross-guided learning for few-shot segmentation,'' in \emph{Proceedings of the IEEE/CVF Conference on Computer Vision and Pattern Recognition}, 2021, pp. 8312--8321.

\bibitem{liu2022learning}
Y.~Liu, N.~Liu, Q.~Cao, X.~Yao, J.~Han, and L.~Shao, ``Learning non-target knowledge for few-shot semantic segmentation,'' in \emph{Proceedings of the IEEE/CVF Conference on Computer Vision and Pattern Recognition}, 2022, pp. 11\,573--11\,582.

\bibitem{wang2019panet}
K.~Wang, J.~H. Liew, Y.~Zou, D.~Zhou, and J.~Feng, ``{PANet}: Few-shot image semantic segmentation with prototype alignment,'' in \emph{Proceedings of the IEEE/CVF International Conference on Computer Vision}, 2019, pp. 9196--9205.

\bibitem{xie2021few}
G.~Xie, H.~Xiong, J.~Liu, Y.~Yao, and L.~Shao, ``Few-shot semantic segmentation with cyclic memory network,'' in \emph{Proceedings of the IEEE/CVF International Conference on Computer Vision}, 2021, pp. 7273--7282.

\bibitem{xie2021scale}
G.~Xie, J.~Liu, H.~Xiong, and L.~Shao, ``Scale-aware graph neural network for few-shot semantic segmentation,'' in \emph{Proceedings of the IEEE Conference on Computer Vision and Pattern Recognition}, 2021, pp. 5475--5484.

\bibitem{DBLP:journals/tmm/ChenXYWSTZ22}
T.~Chen, G.~Xie, Y.~Yao, Q.~Wang, F.~Shen, Z.~Tang, and J.~Zhang, ``Semantically meaningful class prototype learning for one-shot image segmentation,'' \emph{{IEEE} Trans. Multim.}, vol.~24, pp. 968--980, 2022.

\bibitem{zhang2021rich}
X.~Zhang, Y.~Wei, Z.~Li, C.~Yan, and Y.~Yang, ``Rich embedding features for one-shot semantic segmentation,'' \emph{IEEE Transactions on Neural Networks and Learning Systems}, vol.~33, no.~11, pp. 6484--6493, 2021.

\bibitem{dong2018few}
N.~Dong and E.~P. Xing, ``Few-shot semantic segmentation with prototype learning,'' in \emph{British Machine Vision Conference}, 2018, pp. 1--13.

\bibitem{lang2022beyond}
C.~Lang, B.~Tu, G.~Cheng, and J.~Han, ``Beyond the prototype: Divide-and-conquer proxies for few-shot segmentation,'' in \emph{Proceedings of the International Joint Conference on Artificial Intelligence}, 2022, pp. 1024--1030.

\bibitem{long2015fully}
J.~Long, E.~Shelhamer, and T.~Darrell, ``Fully convolutional networks for semantic segmentation,'' in \emph{Proceedings of the IEEE Conference on Computer Vision and Pattern Recognition}, 2015, pp. 3431--3440.

\bibitem{chen2018encoder}
L.~Chen, Y.~Zhu, G.~Papandreou, F.~Schroff, and H.~Adam, ``Encoder-decoder with atrous separable convolution for semantic image segmentation,'' in \emph{Proceedings of the European Conference on Computer Vision}, vol. 11211, 2018, pp. 833--851.

\bibitem{zhao2017pyramid}
H.~Zhao, J.~Shi, X.~Qi, X.~Wang, and J.~Jia, ``Pyramid scene parsing network,'' in \emph{Proceedings of the IEEE Conference on Computer Vision and Pattern Recognition}, 2017, pp. 6230--6239.

\bibitem{he2019adaptive}
J.~He, Z.~Deng, L.~Zhou, Y.~Wang, and Y.~Qiao, ``Adaptive pyramid context network for semantic segmentation,'' in \emph{Proceedings of the IEEE/CVF Conference on Computer Vision and Pattern Recognition}, 2019, pp. 7519--7528.

\bibitem{crm/tip22/CIRNet}
R.~Cong, Q.~Lin, C.~Zhang, C.~Li, X.~Cao, Q.~Huang, and Y.~Zhao, ``{CIR-Net}: Cross-modality interaction and refinement for {RGB-D} salient object detection,'' \emph{IEEE Trans. Image Process.}, vol.~31, pp. 6800--6815, 2022.

\bibitem{chen2014semantic}
L.~Chen, G.~Papandreou, I.~Kokkinos, K.~Murphy, and A.~L. Yuille, ``Semantic image segmentation with deep convolutional nets and fully connected {CRFs},'' in \emph{International Conference on Learning Representations}, 2015.

\bibitem{wang2018non}
X.~Wang, R.~B. Girshick, A.~Gupta, and K.~He, ``Non-local neural networks,'' in \emph{Proceedings of the IEEE Conference on Computer Vision and Pattern Recognition}, 2018, pp. 7794--7803.

\bibitem{huang2019ccnet}
Z.~Huang, X.~Wang, L.~Huang, C.~Huang, Y.~Wei, and W.~Liu, ``{CCNet}: Criss-cross attention for semantic segmentation,'' in \emph{Proceedings of the IEEE/CVF International Conference on Computer Vision}, 2019, pp. 603--612.

\bibitem{zhang2020dynamic}
L.~Zhang, D.~Xu, A.~Arnab, and P.~H.~S. Torr, ``Dynamic graph message passing networks,'' in \emph{Proceedings of the IEEE/CVF Conference on Computer Vision and Pattern Recognition}, 2020, pp. 3723--3732.

\bibitem{cheng2021per}
B.~Cheng, A.~G. Schwing, and A.~Kirillov, ``Per-pixel classification is not all you need for semantic segmentation,'' in \emph{Advances in Neural Information Processing Systems}, 2021, pp. 17\,864--17\,875.

\bibitem{cheng2022masked}
B.~Cheng, I.~Misra, A.~G. Schwing, A.~Kirillov, and R.~Girdhar, ``Masked-attention mask transformer for universal image segmentation,'' in \emph{Proceedings of the IEEE/CVF Conference on Computer Vision and Pattern Recognition}, 2022, pp. 1280--1289.

\bibitem{wang2020generalizing}
Y.~Wang, Q.~Yao, J.~T. Kwok, and L.~M. Ni, ``Generalizing from a few examples: {A} survey on few-shot learning,'' \emph{{ACM} Comput. Surv.}, vol.~53, no.~3, pp. 63:1--63:34, 2021.

\bibitem{ravi2017optimization}
S.~Ravi and H.~Larochelle, ``Optimization as a model for few-shot learning,'' in \emph{International Conference on Learning Representations}, 2017.

\bibitem{sung2018learning}
F.~Sung, Y.~Yang, L.~Zhang, T.~Xiang, P.~H.~S. Torr, and T.~M. Hospedales, ``Learning to compare: Relation network for few-shot learning,'' in \emph{Proceedings of the IEEE Conference on Computer Vision and Pattern Recognition}, 2018, pp. 1199--1208.

\bibitem{snell2017prototypical}
J.~Snell, K.~Swersky, and R.~S. Zemel, ``Prototypical networks for few-shot learning,'' in \emph{Advances in Neural Information Processing Systems}, 2017, pp. 4077--4087.

\bibitem{sun2019meta}
Q.~Sun, Y.~Liu, T.~Chua, and B.~Schiele, ``Meta-transfer learning for few-shot learning,'' in \emph{Proceedings of the IEEE Conference on Computer Vision and Pattern Recognition}, 2019, pp. 403--412.

\bibitem{oreshkin2018tadam}
B.~N. Oreshkin, P.~R. L{\'{o}}pez, and A.~Lacoste, ``{TADAM}: Task dependent adaptive metric for improved few-shot learning,'' in \emph{Advances in Neural Information Processing Systems}, 2018, pp. 719--729.

\bibitem{allen2019infinite}
K.~R. Allen, E.~Shelhamer, H.~Shin, and J.~B. Tenenbaum, ``Infinite mixture prototypes for few-shot learning,'' in \emph{Proceedings of the IEEE International Conference on Machine Learning}, vol.~97, 2019, pp. 232--241.

\bibitem{koch2015siamese}
G.~Koch, R.~Zemel, R.~Salakhutdinov \emph{et~al.}, ``Siamese neural networks for one-shot image recognition,'' in \emph{Proceedings of the IEEE International Conference on Machine Learning Workshop}, 2015.

\bibitem{jamal2019task}
M.~A. Jamal and G.~Qi, ``Task agnostic meta-learning for few-shot learning,'' in \emph{Proceedings of the IEEE Conference on Computer Vision and Pattern Recognition}, 2019, pp. 11\,719--11\,727.

\bibitem{finn2017model}
C.~Finn, P.~Abbeel, and S.~Levine, ``Model-agnostic meta-learning for fast adaptation of deep networks,'' in \emph{Proceedings of the IEEE International Conference on Machine Learning}, vol.~70, 2017, pp. 1126--1135.

\bibitem{gordon2018meta}
J.~Gordon, J.~Bronskill, M.~Bauer, S.~Nowozin, and R.~E. Turner, ``Meta-learning probabilistic inference for prediction,'' in \emph{International Conference on Learning Representations}, 2019.

\bibitem{grant2018recasting}
E.~Grant, C.~Finn, S.~Levine, T.~Darrell, and T.~L. Griffiths, ``Recasting gradient-based meta-learning as hierarchical bayes,'' in \emph{International Conference on Learning Representations}, 2018.

\bibitem{chen2019image}
Z.~Chen, Y.~Fu, K.~Chen, and Y.~Jiang, ``Image block augmentation for one-shot learning,'' in \emph{Proceedings of the AAAI Conference on Artificial Intelligence}, 2019, pp. 3379--3386.

\bibitem{shaban2017one}
A.~Shaban, S.~Bansal, Z.~Liu, I.~Essa, and B.~Boots, ``One-shot learning for semantic segmentation,'' in \emph{British Machine Vision Conference}, 2017, pp. 1--13.

\bibitem{zhang2019canet}
C.~Zhang, G.~Lin, F.~Liu, R.~Yao, and C.~Shen, ``{CANet}: Class-agnostic segmentation networks with iterative refinement and attentive few-shot learning,'' in \emph{Proceedings of the IEEE/CVF Conference on Computer Vision and Pattern Recognition}, 2019, pp. 5217--5226.

\bibitem{boudiaf2021few}
M.~Boudiaf, H.~Kervadec, I.~M. Ziko, P.~Piantanida, I.~B. Ayed, and J.~Dolz, ``Few-shot segmentation without meta-learning: {A} good transductive inference is all you need?'' in \emph{Proceedings of the IEEE/CVF Conference on Computer Vision and Pattern Recognition}, 2021, pp. 13\,979--13\,988.

\bibitem{zhang2021few}
G.~Zhang, G.~Kang, Y.~Yang, and Y.~Wei, ``Few-shot segmentation via cycle-consistent transformer,'' in \emph{Advances in Neural Information Processing Systems}, vol.~34, 2021, pp. 21\,984--21\,996.

\bibitem{mao2022learning}
B.~Mao, X.~Zhang, L.~Wang, Q.~Zhang, S.~Xiang, and C.~Pan, ``Learning from the target: Dual prototype network for few shot semantic segmentation,'' in \emph{Proceedings of the AAAI Conference on Artificial Intelligence}, vol.~36, no.~2, 2022, pp. 1953--1961.

\bibitem{wang2020few}
H.~Wang, X.~Zhang, Y.~Hu, Y.~Yang, X.~Cao, and X.~Zhen, ``Few-shot semantic segmentation with democratic attention networks,'' in \emph{Proceedings of the European Conference on Computer Vision}, vol. 12358, 2020, pp. 730--746.

\bibitem{hou2019cross}
R.~Hou, H.~Chang, B.~Ma, S.~Shan, and X.~Chen, ``Cross attention network for few-shot classification,'' in \emph{Advances in Neural Information Processing Systems}, 2019, pp. 4005--4016.

\bibitem{fu2019dual}
J.~Fu, J.~Liu, H.~Tian, Y.~Li, Y.~Bao, Z.~Fang, and H.~Lu, ``Dual attention network for scene segmentation,'' in \emph{Proceedings of the IEEE Conference on Computer Vision and Pattern Recognition}, 2019, pp. 3146--3154.

\bibitem{zhu2019asymmetric}
Z.~Zhu, M.~Xu, S.~Bai, T.~Huang, and X.~Bai, ``Asymmetric non-local neural networks for semantic segmentation,'' in \emph{Proceedings of the IEEE/CVF International Conference on Computer Vision}, 2019, pp. 593--602.

\bibitem{nguyen2019feature}
K.~Nguyen and S.~Todorovic, ``Feature weighting and boosting for few-shot segmentation,'' in \emph{Proceedings of the IEEE/CVF International Conference on Computer Vision}, 2019, pp. 622--631.

\bibitem{everingham2010pascal}
M.~Everingham, L.~V. Gool, C.~K.~I. Williams, J.~M. Winn, and A.~Zisserman, ``The pascal visual object classes {(VOC)} challenge,'' \emph{Int. J. Comput. Vis.}, vol.~88, no.~2, pp. 303--338, 2010.

\bibitem{hariharan2011semantic}
B.~Hariharan, P.~Arbelaez, L.~D. Bourdev, S.~Maji, and J.~Malik, ``Semantic contours from inverse detectors,'' in \emph{Proceedings of the IEEE/CVF International Conference on Computer Vision}, 2011, pp. 991--998.

\bibitem{lin2014microsoft}
T.~Lin, M.~Maire, S.~J. Belongie, J.~Hays, P.~Perona, D.~Ramanan, P.~Doll{\'{a}}r, and C.~L. Zitnick, ``Microsoft {COCO}: Common objects in context,'' in \emph{Proceedings of the European Conference on Computer Vision}, 2014, pp. 740--755.

\bibitem{yang2021mining}
L.~Yang, W.~Zhuo, L.~Qi, Y.~Shi, and Y.~Gao, ``Mining latent classes for few-shot segmentation,'' in \emph{Proceedings of the IEEE/CVF International Conference on Computer Vision}, 2021, pp. 8701--8710.

\bibitem{liu2022dynamic}
J.~Liu, Y.~Bao, G.~Xie, H.~Xiong, J.~Sonke, and E.~Gavves, ``Dynamic prototype convolution network for few-shot semantic segmentation,'' in \emph{Proceedings of the IEEE/CVF Conference on Computer Vision and Pattern Recognition}, 2022, pp. 11\,553--11\,562.

\bibitem{paszke2019pytorch}
A.~Paszke, S.~Gross, F.~Massa, A.~Lerer, J.~Bradbury, G.~Chanan, T.~Killeen, Z.~Lin, N.~Gimelshein, L.~Antiga, A.~Desmaison, A.~K{\"{o}}pf, E.~Z. Yang, Z.~DeVito, M.~Raison, A.~Tejani, S.~Chilamkurthy, B.~Steiner, L.~Fang, J.~Bai, and S.~Chintala, ``{PyTorch}: An imperative style, high-performance deep learning library,'' in \emph{Advances in Neural Information Processing Systems}, 2019, pp. 8024--8035.

\bibitem{fan2022self}
Q.~Fan, W.~Pei, Y.~Tai, and C.~Tang, ``Self-support few-shot semantic segmentation,'' in \emph{Proceedings of the European Conference on Computer Vision}, vol. 13679, 2022, pp. 701--719.

\bibitem{he2016deep}
K.~He, X.~Zhang, S.~Ren, and J.~Sun, ``Deep residual learning for image recognition,'' in \emph{Proceedings of the IEEE Conference on Computer Vision and Pattern Recognition}, 2016, pp. 770--778.

\bibitem{russakovsky2015imagenet}
O.~Russakovsky, J.~Deng, H.~Su, J.~Krause, S.~Satheesh, S.~Ma, Z.~Huang, A.~Karpathy, A.~Khosla, M.~S. Bernstein, A.~C. Berg, and L.~Fei{-}Fei, ``Imagenet large scale visual recognition challenge,'' \emph{Int. J. Comput. Vis.}, vol. 115, no.~3, pp. 211--252, 2015.

\bibitem{liu2022intermediate}
Y.~Liu, N.~Liu, X.~Yao, and J.~Han, ``Intermediate prototype mining transformer for few-shot semantic segmentation,'' in \emph{Advances in Neural Information Processing Systems}, 2022, pp. 38\,020--38\,031.

\end{thebibliography}

\end{document}